%% file: main.tex
\documentclass{article}

\PassOptionsToPackage{numbers, compress}{natbib}

\usepackage[final]{neurips_2021}

\usepackage[utf8]{inputenc} %
\usepackage[T1]{fontenc}    %
\usepackage{hyperref}       %
\usepackage{url}            %
\usepackage{booktabs}       %
\usepackage{amsfonts}       %
\usepackage{nicefrac}       %
\usepackage{microtype}      %
\usepackage{xcolor}         %
\usepackage{graphicx}
\usepackage{amsmath}
\usepackage{amssymb}
\usepackage{bbm}
\usepackage{multirow}
\usepackage{pifont}
\usepackage{algorithm,algorithmicx,algpseudocode}
\usepackage{wrapfig,lipsum,booktabs}
\usepackage{caption}
\usepackage{etoc}
\usepackage{appendix}
\usepackage[symbol]{footmisc}

\DeclareCaptionLabelFormat{andtable}{#1~#2  \&  \tablename~\thetable}

\newcommand{\bs}[1]{\boldsymbol{#1}}
\newcommand{\cmark}{\ding{51}}%
\newcommand{\xmark}{\ding{55}}%

\hypersetup{
colorlinks=true,
linkcolor=blue,
citecolor=blue,
filecolor=magenta,
urlcolor=blue
}

\title{Dynamics-Regulated Kinematic Policy for\\Egocentric Pose Estimation}

\author{%
  Zhengyi Luo\textsuperscript{\rm 1} \quad  Ryo  Hachiuma \textsuperscript{\rm 2} \thanks{Work done at Carnegie Mellon University.} \quad Ye Yuan\textsuperscript{\rm 1} \quad Kris Kitani\textsuperscript{\rm 1} \\
    \textsuperscript{\rm 1} Carnegie Mellon University \quad  \textsuperscript{\rm 2} Keio University\\
    \url{https://zhengyiluo.github.io/projects/kin\_poly/} 
}

\begin{document}
\maketitle

\begin{abstract}
We propose a method for object-aware 3D egocentric pose estimation that tightly integrates kinematics modeling, dynamics modeling, and scene object information. Unlike prior kinematics or dynamics-based approaches where the two components are used disjointly, we synergize the two approaches via \emph{dynamics-regulated training}. At each timestep, a kinematic model is used to provide a target pose using video evidence and simulation state. Then, a prelearned dynamics model attempts to mimic the kinematic pose in a physics simulator. By comparing the pose instructed by the kinematic model against the pose generated by the dynamics model, we can use their misalignment to further improve the kinematic model. By factoring in the 6DoF pose of objects (e.g., chairs, boxes) in the scene, we demonstrate for the first time, the ability to estimate physically-plausible 3D human-object interactions using a single wearable camera. We evaluate our egocentric pose estimation method in both controlled laboratory settings and real-world scenarios.
\end{abstract}

\etocdepthtag.toc{mtchapter}
\etocsettagdepth{mtchapter}{subsection}
\etocsettagdepth{mtappendix}{none}

\vspace{-3mm}
\section{Introduction}
\vspace{-2mm}

From a video captured by a single head-mounted wearable camera (\eg, smartglasses, action camera, body camera), we aim to infer the wearer's global 3D full-body pose and interaction with objects in the scene, as illustrated in Fig. \ref{fig:teaser}. This is important for applications like virtual and augmented reality, sports analysis, and wearable medical monitoring, where third-person views are often unavailable and proprioception algorithms are needed for understanding the actions of the camera wearer. However, this task is challenging since the wearer's body is often unseen from a first-person view and the body motion needs to be inferred solely based on the videos captured by the \emph{front-facing} camera. Furthermore, egocentric videos usually capture the camera wearer interacting with objects in the scene, which adds additional complexity in recovering a pose sequence that agrees with the scene context. Despite these challenges, we show that it is possible to infer accurate human motion and human-object interaction from a single head-worn front-facing camera. 

Egocentric pose estimation can be solved using two different paradigms: (1) a kinematics perspective and (2) a dynamics perspective. \emph{Kinematics-based approaches} study motion without regard to the underlying forces (\eg, gravity, joint torque) and cannot faithfully emulate human-object interaction without modeling proper contact and forces. They can achieve accurate pose estimates by directly outputting joint angles but can also produce results that violate physical constraints (\eg foot skating and ground penetration). \emph{Dynamics-based approaches}, or \emph{physics-based approaches}, study motions that result from forces. They map directly from visual input to control signals of a human proxy (humanoid) inside a physics simulator and recover 3D poses through simulation. These approaches have the crucial advantage that they output physically-plausible human motion and human-object interaction (\ie, pushing an object will move it according to the rules of physics). However, since no joint torque is captured in human motion datasets, physics-based humanoid controllers are hard to learn, generalize poorly, and are actively being researched \cite{Peng2018DeepMimic, yuan2020residual, Wang2020UniConUN, ScaDiver}.

In this work, we argue that a \emph{hybrid} approach merging the kinematics and dynamics perspectives is needed. Leveraging a large human motion database \cite{Mahmood2019AMASSAO}, we learn a task-agnostic dynamics-based humanoid controller to mimic broad human behaviors, ranging from every day motion to dancing and kickboxing. The controller is \emph{general-purpose} and can be viewed as providing low-level motor skills of a human. After the controller is learned, we train an object-aware kinematic policy to specify the target poses for the controller to mimic. One approach is to let the kinematic model produce target motion \emph{only} based on the visual input \cite{yuan2021simpoe, Wang2020UniConUN, yuan2019ego}. This approach uses the physics simulation as a post-processing step: the kinematic model computes the target motion separately from the simulation and may output unreasonable target poses. We propose to synchronize the two aspects by designing a kinematic policy that guides the controller and receives timely feedback through comparing its target pose and the resulting simulation state. Our model thus serves as a high-level motion planning module that adapts intelligently based on the current simulation state. In addition, since our kinematic policy only outputs poses and does not model joint torque, it can receive direct supervision from  motion capture (MoCap) data. While poses from MoCap can provide an initial-guess of target motion, our model can search for better solutions through trial and error. This learning process, dubbed \emph{dynamics-regulated training}, jointly optimizes our model via supervised learning and reinforcement learning, and significantly improves its robustness to real-world use cases.

In summary, our contributions are as follows: (1) we are the first to tackle the challenging task of estimating physically-plausible 3D poses and human-object interactions from a single front-facing camera; (2) we learn a general-purpose humanoid controller from a large MoCap dataset and can perform a broad range of motions inside a physics simulation; (3) we propose a dynamics-regulated training procedure that synergizes kinematics, dynamics, and scene context for egocentric vision; (4)~experiments on a controlled motion capture laboratory dataset and a real-world dataset demonstrate that our model outperforms other state-of-the-art methods on pose-based and physics-based metrics, while generalizing to videos taken in real-world scenarios.

\vspace{-3mm}
\section{Related Work}
\vspace{-2mm}

\noindent \textbf{Third-person human pose estimation.} The task of estimating the 3D human pose (and sometimes shape) from \textit{third-person} video is a popular research area in the vision community \cite{Bogo2016KeepIS, Kanazawa2018EndtoEndRO, Kolotouros2019LearningTR, Hossain2018, Rogez2019LCR, Zhang2019, Georgakis, Pavllo20193D,Habibie2019In,Moon2019MPPE, Kocabas2020VIBEVI, Luo20203DHM, Gler2018DensePoseDH, Xu2019DenseRaCJ3, yuan2021simpoe}, with methods aiming to recover 3D joint positions \cite{Li20143DHP, Tekin2016StructuredPO, Hossain2018, Pavllo20193D}, 3D joint angles with respect to a parametric human model \cite{Bogo2016KeepIS, Kanazawa2018EndtoEndRO,  Kocabas2020VIBEVI, Luo20203DHM}, and dense body parts \cite{Gler2018DensePoseDH}. Notice that all these  methods are purely kinematic and disregard physical reasoning. They also do not recover the global 3D root position and are evaluated by zeroing out the body center (root-relative). A smaller number of works factor in human dynamics \cite{PhysAwareTOG2021,Shimada2020PhysCapPP, Rempe2020ContactAH, Vondrak2012Videobased3M, Brubaker2009EstimatingCD} through postprocessing, physics-based trajectory optimization, or using a differentiable physics model. These approaches can produce physically-plausible human motion, but since they do not utilize a physics simulator and does not model contact, they can not faithfully model human-object interaction. SimPoE \cite{yuan2021simpoe}, a recent work on third-person pose estimation using simulated character control, is most related to ours, but 1) trains a single and dataset-specific humanoid controller per dataset; 2) designs the kinematic model to be independent from simulation states.

\noindent \textbf{Egocentric human pose estimation.} Compared to third-person human pose estimation, there are only a handful of attempts at estimating 3D full body poses from  egocentric videos due to the ill-posed nature of this task. Most existing methods still assume partial visibility of body parts in the image \cite{tome2019xr,Rhodin2016EgoCap,xu2019mo2cap2}, often through a downward-facing camera. Among works where the human body is mostly not observable  \cite{Jiang2016SeeingIP,Yuan_2018_ECCV,yuan2019ego,Ng2019You}, Jiang \etal \cite{Jiang2016SeeingIP} use a kinematics-based approach where they construct a motion graph from the training data and recover the pose sequence by solving the optimal pose path. Ng \etal \cite{Ng2019You} focus on modeling person-to-person interactions from egocentric videos and inferring the wearer's pose conditioning on the other person's pose. The works most related to ours are \cite{Yuan_2018_ECCV, yuan2019ego,isogawa2020optical} which use dynamics-based approaches and map visual inputs to control signals to perform physically-plausible human motion inside a physics simulation. They show impressive results on a set of noninteractive locomotion tasks, but also observe large errors in absolute 3D position tracking--mapping directly from the visual inputs to control signals is a noisy process and prone to error accumulation. In comparison, our work jointly models kinematics and dynamics, and estimates a wider range of human motion and human-object interactions while improving absolute 3D position tracking. To the best of our knowledge, we are the first approach to estimate the 3D human poses from egocentric video while factoring in human-object interactions.

\noindent \textbf{Humanoid control inside physics simulation.} 
Our work is also connected to controlling humanoids to mimic reference motion \cite{Peng2018DeepMimic,Peng2018SFV, chao2019learning, yuan2020residual, ScaDiver, Soohwan2019learning, Chentanez2018PhysicsbasedMC} and interact with objects \cite{chao2019learning,merel2019reusable} inside a physics simulator. The core motivation of these works is to learn the necessary dynamics to imitate or generate human motion in a physics simulation. Deep RL has been the predominant approach in this line of work since physics simulators are typically not end-to-end differetiable. Goal-oriented methods \cite{chao2019learning, merel2019reusable, Bergamin2019DReCon} does not involve motion imitation and are evaluated on task completion (moving an object, sitting on a chair, moving based on user-input \etc). Consequently, these frameworks only need to master a subset of possible motions for task completion. People, on the other hand, have a variety of ways to perform actions, and our agent has to follow the trajectory predefined by egocentric videos. Motion imitation methods \cite{Peng2018DeepMimic, Peng2018SFV, ScaDiver, yuan2020residual, ScaDiver} aim to control characters to mimic a sequence of reference motion, but have been limited to performing a single clip \cite{Peng2018DeepMimic, Peng2018SFV, ScaDiver, yuan2020residual} or high-quality MoCap \cite{ScaDiver} motion (and requires fine-tuning to generalize to other motion generators). In contrast, our dynamics controller is general and can be used to perform everyday motion and human-object interactions estimated by a kinematic motion estimator without task-specific fine-tuning.

\begin{figure*}[t]
\begin{center}
\includegraphics[width=1\hsize]{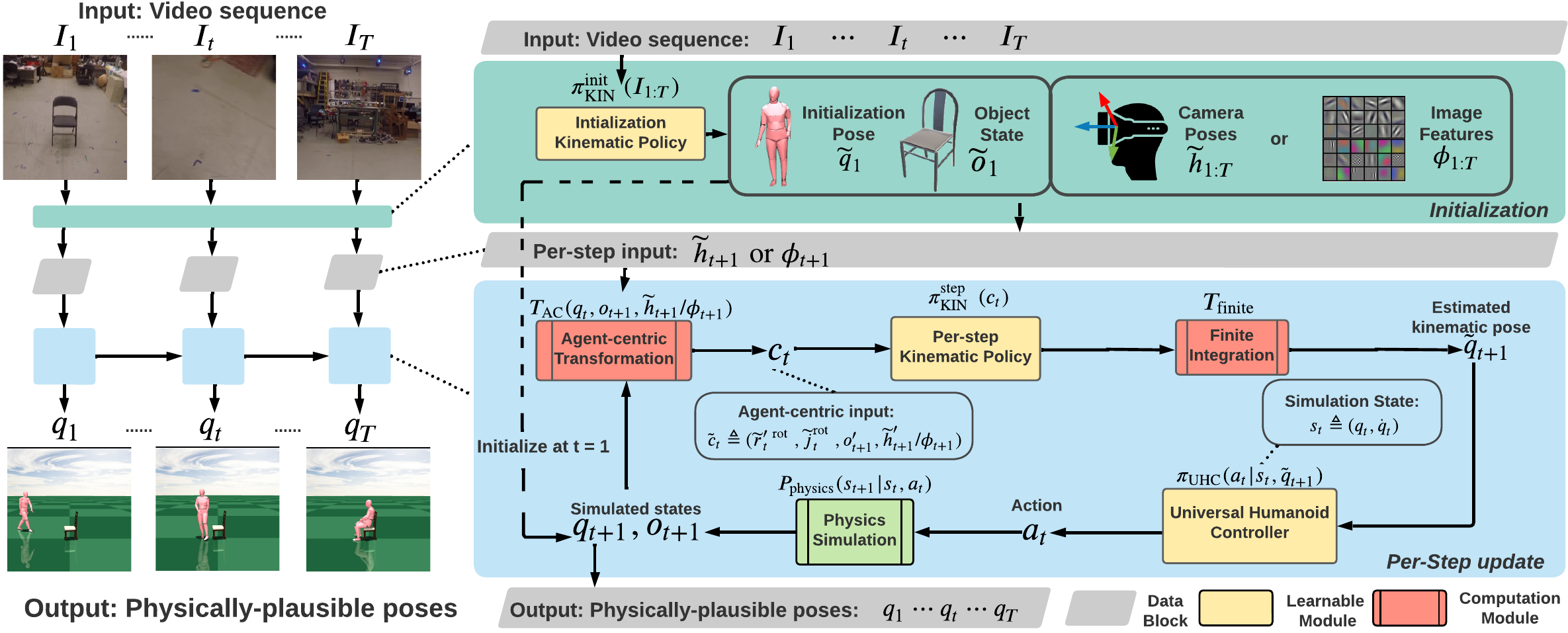}
\end{center}
\vspace{-2mm}
\caption{Overview of our dynamics-regulated kinematic policy. Given an egocentirc video $\boldsymbol{I}_{1: T}$, our initialization module $\boldsymbol{\pi}_{\text{KIN}}^{\text {init }}\left(\boldsymbol{I}_{1: T}\right)$ computes the first-frame object state $\widetilde{\bs o}_{1}$, human pose $\widetilde{\bs q}_{1}$, camera poses $\widetilde{\bs h}_{1:T}$ or image features $\bs \phi_{1:T}$. The object state $\widetilde{\bs o}_{1}$ and human pose $\widetilde{\bs q}_{1}$ are used to initialize the phsycis simulation. At each time step, we roll out our per-step kinematic policy ${\bs\pi}_{\text{KIN}}^{\text{step}}$ together with the Universal Humanoid Controller to output  physically-plausible pose $\bs q_t$ inside a physics simulator. 
 }
\vspace{-3mm}
\label{fig:overview}
\end{figure*}

\vspace{-3mm}
\section{Method}
\vspace{-2mm}

The problem of egocentric pose estimation can be formulated as follows: from a wearable camera footage $\bs{I}_{1:T}$, we want to recover the wearer's ground truth global 3D poses ${\widehat{\bs{q}}_{1:T}}$. Each pose $\widehat{\bs{q}}_t  \triangleq (\widehat{\bs{r}}^{\text{pos}}_t, \widehat{\bs{r}}^{\text{rot}}_t, \widehat{\bs{j}}^{\text{rot}}_t)$ consists of the root position $\widehat{\bs{r}}^{\text{pos}}_t$, root orientation $\widehat{\bs{r}}^{\text{rot}}_t$ , and body joint angles $\widehat{\bs{j}}^{\text{rot}}_t$ of the human model. Here we adopt the popular SMPL \cite{Loper2015SMPLAS} human model and the humanoid we use in physics simulation is created from the kinematic structure and mean body shape defined by SMPL.
Our framework first learns a Universal Humanoid Controller (UHC) from a large MoCap dataset (Sec. \ref{sec:UHC}). The learned UHC can be viewed as providing the lower level muscle skills of a real human, trained by mimicking thousands of human motion sequences. Using the trained UHC, we learn our kinematic policy (Sec. \ref{sec:kin_policy}) through dynamics-regulated training (Sec. \ref{sec:dynamics_regulated}). At the test time, the kinematic policy provides per-step target motion to the UHC, forming a closed-loop system that operates inside the physics simulation to control a humanoid. The result of the UHC and physics simulation is then used as input to the kinematic policy to produce the next-frame target motion, as depicted in Fig. \ref{fig:overview}. As a notation convention, we use $\widetilde{\cdot}$ to denote kinematic quantities (obtained without using physics simulation), $\widehat{\cdot}$ to denote ground truth quantities, and normal symbols without accents to denote quantities from the physics simulation.

\vspace{-3mm}
\subsection{Dynamics Model - Universal Humanoid Controller (UHC)}
\label{sec:UHC}
\vspace{-2mm}
To learn a task-agnostic dynamics model that can be tightly integrated with a kinematic model, we design our controller's state space to only rely on the current simulated pose ${\bs{q}}_{t}$ and target posed $\widehat{\bs{q}}_{t+1}$ and remove all phase or sequence-level information found in prior arts \cite{yuan2020residual, Peng2018DeepMimic, Peng2018SFV}. This design allows us to train on an order of magnitude larger dataset of human motion \cite{Mahmood2019AMASSAO} with only pose information and significantly improve our models' ability to mimic diverse and unseen motions. Formally, we model controlling a humanoid to follow a reference motion $\widehat{\bs{q}}_{1:T}$ as a Markov Decision Process (MDP) defined as a tuple ${\mathcal M}=\langle S, A, P_{\text{physics}}, R, \gamma\rangle$ of states, actions, transition dynamics, reward function, and a discount factor. The state $S$, reward $R$, and transition dynamics $P_{\text{physics}}$ are provided by the physics simulator, while action $A$ is computed by the policy ${\bs \pi}_{\text{UHC}}$. At each timestep $t$, the agent in state $\bs{s}_t$ takes an action sampled from the policy ${\bs \pi}_{\text{UHC}}(\bs{a}_t|\bs{s}_t, \widehat{\bs{q}}_{t+1})$ while the environment generates the next state $s_{t+1}$ and reward $r_t$. We employ Proximal Policy Optimization (PPO) \cite{schulman2017proximal} to find the optimal policy ${\bs \pi}_{\text{UHC}}^\ast$ that maximizes the expected discounted return $\mathbb{E}[\sum_{t=1}^T \gamma^{t-1}r_t]$. 

\noindent \textbf{State.} The state $\bs s_{t} \triangleq\left({\bs{q}}_{t}, \dot{\bs{q}}_{t} \right)$ of the humanoid contains the character's current pose ${\bs{q}}_{t}$ and joint velocity $\dot{\bs{q}}_{t}$. Here, the state $\bs s_{t}$ encapsulates the humanoid's full physical state at time step $t$. It only includes information about the current frame $({\bs{q}}_{t}, \dot{\bs{q}}_{t})$ and does not include any extra information, enabling our learned controller to be guided by a target pose only.

\noindent \textbf{Action.} The action $\bs{a}_t$ specifies the target joint angles for the proportional derivative (PD) controller \cite{Tan2011Stable} at each degree of freedom (DoF) of the humanoid joints except for the root (pelvis). We use the residual action representation \cite{Soohwan2019learning}: $\bs{q}^d_t = \widehat{\bs{q}}_{t+1} + \bs a_t$, 
 where $\bs{q}^d_t$ is the final PD target, $\bs a_t$ is the output of the control policy ${\bs \pi}_{\text{UHC}}$, and $\widehat{\bs{q}}_{t+1}$ is the target pose. The torque to be applied at joint $i$ is: ${\bs \tau}^i = {\bs k}^{p} \circ (\bs{q}^d_t-\bs{q}_t)- \bs{k}^d \circ \dot{\bs{q}}_{t}$ where $\bs{k}^p$ and  $\bs{k}^d$ are manually specified gains and $\circ$ is the element-wise multiplication. As observed in prior work \cite{yuan2020residual, yuan2021simpoe}, allowing the policy to apply external residual forces $\bs{\eta}_t$ to the root helps stabilizing the humanoid, so our final action is ${\bs{a}}_{t} \triangleq(\Delta \widetilde{\bs{q}}^d_t, \bs{\eta}_{t})$.

\noindent \textbf{Policy.} The policy $\bs {\bs \pi}_{\text{UHC}}(\bs{a}_t|\bs{s}_t, \widehat{\bs{q}}_{t+1})$ is represented by a Gaussian distribution with a fixed diagonal covariance matrix $\Sigma$. We first use a feature extraction layer $D_{\text{diff}}(\widehat{\bs{q}}_{t+1}, {\bs{q}}_{t})$ to compute the root and joint offset between the simulated pose and target pose. All features are then translated to a root-relative coordinate system using an agent-centic transform ${\bs T}_{\text{AC}}$ to make our policy orientation-invariant. We use a Multi-layer Perceptron (MLP) as our policy network to map the augmented state ${\bs T}_{\text{AC}}\left({\bs{q}}_{t}, \dot{\bs{q}}_{t}, \widehat{\bs{q}}_{t+1}, D_{\text{diff}}(\widehat{\bs{q}}_{t+1}, {\bs{q}}_{t}) \right)$ to the predicted action $\bs{a}_t$. 

\noindent \textbf{Reward function.} For UHC, the reward function is designed to encourage the simulated pose ${\bs{q}}_t$ to better match the target pose $\widehat{\bs{q}}_{t+1}$. Since we share a similar objective (mimic target motion), our reward is similar to Residual Force Control \cite{yuan2020residual}.

\noindent \textbf{Training procedure.} We train our controller on the AMASS \cite{Mahmood2019AMASSAO} dataset, which contains 11505 high-quality MoCap sequences with 4000k frame of poses (after removing sequences involving human-object interaction like running on a treadmill). At the beginning of each episode, a random fixed length sequence (300 frames) is sampled from the dataset for training. While prior works \cite{ScaDiver, Wang2020UniConUN} uses more complex motion clustering techniques to sample motions, we devise a simple yet empirically effective sampling technique by inducing a probability distribution based on the value function. For each pose frame $\widehat{\bs{q}}_{j}$ in the dataset, we first compute an initialization state $\bs{s}^{j}_{1}$: $\bs{s}^{j}_{1} \triangleq \left(\widehat{\bs{q}}_{j}, \bs 0 \right)$, and then score it using the value function to access how well the policy can mimic the sequence $\widehat{\bs{q}}_{j:T}$ starting from this pose: $V(\bs{s}^{j}_{1}, \widehat{\bs{q}}_{j+1}) =  v_j$. Intuitively, the higher $\bs v_j$ is, the more confident our policy is in mimicing this sequence, and the less often we should pick this frame. The probability of choosing frame $j$, comparing against all frames $J$ in the AMASS dataset, is then  $P(\widehat{\bs{q}}_{j}) = \frac{\exp(-v_j/\tau)}{\sum_i^J \exp(-v_i/\tau)}$ where $\tau$ is the sampling temperature. 
More implementation details about the reward, training and evaluation of UHC can be found in Appendix \ref{app:uhc}.

\subsection{Kinematic Model -- Object-aware Kinematic Policy}
\label{sec:kin_policy}
To leverage the power of our learned UHC, we design an auto-regressive and object-aware kinematic policy to generate per-frame target motion from egocentric inputs. We synchronize the state space of our kinematic policy and UHC such that the policy can be learned with or without physics simulation. When trained without physics simulation, the model is purely kinematic and can be optimized via supervised learning; when trained with a physics simulation, the model can be optimized through a combination of supervised learning and reinforcement learning. The latter procedure, coined \textit{dynamics-regulated training}, enables our model to distill human dynamics information learned from large-scale MoCap data into the kinematic model and learns a policy more robust to convariate shifts. In this section, we will describe the architecture of the policy itself and the training through supervised learning (without physics simulation). 

\noindent \textbf{Scene context modelling and initialization.} To serve as a high-level target motion estimator for egocentric videos with potential human-object interaction, our kinematic policy needs to be object-aware and grounded with visual input. To this end, given an input image sequence $\bs{I}_{1:T}$, we compute the initial object states $\widetilde{\bs o}_{1}$, camera trajectory $\widetilde{\bs h}_{1:T}$ or image features $\bs{\phi}_{1:T}$ as inputs to our system. The object states, $\widetilde{\bs o_t} \triangleq (\widetilde{\bs{o}}^{cls}_t, \widetilde{\bs{o}}^{\text{pos}}_t, \widetilde{\bs{o}}^{\text{rot}}_t)$, is modeled as a vector concatenation of the main object-of-interest's class $\widetilde{\bs{o}}^{cls}_t$, 3D position $\widetilde{\bs{o}}^{\text{pos}}_t$, and rotation $\widetilde{\bs{o}}^{\text{rot}}_t$. $\widetilde{\bs o_t}$ is computed using an off-the-shelf object detector and pose estimator \cite{ARKitPose}.  When there are no objects in the current scene (for walking and running \etc), the object states vector is set to zero. To provide our model with movement cues, we can either use image level optical flow feature ${\bs \phi}_{1:T}$ or camera trajectory extracted from images using SLAM or VIO. Concretely, the image features ${\bs \phi}_{1:T}$ is computed using an optical flow extractor \cite{Sun2018PWC} and ResNet \cite{He2016DeepRL}. The camera trajectory in our real-world experiments are computed using an off-the-shelf VIO method \cite{ARKitCam}: $\widetilde{\bs h}_t \triangleq (\widetilde{\bs{h}}^{\text{pos}}_t, \widetilde{\bs{h}}^{\text{rot}}_t)$ (position $\widetilde{\bs h}^{\text{pos}}_t$ and orientation $\widetilde{\bs h}^{\text{rot}}_t$).  As visual input can be noisy, using the camera trajectory $\widetilde{\bs h}_t$ directly significantly improves the performance of our framework as shown in our ablation studies (Sec. \ref{sec:abla}). 

To provide our UHC with a plausible initial state for simulation, we estimate $\bs{\widetilde{q}}_1$ from the scene context features $\widetilde{\bs o}_{1:T}$ and $\bs{\phi}_{1:T}$ / $\widetilde{\bs h}_{1:T}$. We use an Gated Recurrent Unit (GRU) \cite{Cho2014LearningPR} based network to regress the initial agent pose $\bs{\widetilde{q}}_1$. Combining the above procedures, we obtain the context modelling and initialization model $\bs{\pi}_{\text{KIN}}^{\text{init}}$: $[\bs{\widetilde{q}}_1, \widetilde{\bs o}_{1},\widetilde{\bs h}_{1:T} / \bs{\phi}_{1:T},] = \bs{\pi}_{\text{KIN}}^{\text{init}}({\bs I_{1:T}})$. Notice that to constrain the ill-posed problem of egocentric pose estimation, we assume known object category, rough size, and potential mode of interaction. We use these knowledge as a prior for our per-step model.

\noindent\textbf{Training kinematic policy via supervised learning.} After initialization, we use the estimated first-frame object pose $\widetilde{\bs o_t}$ and human pose $\widetilde{\bs q_t}$ to \emph{initialize} the physics simulation. All subsequent object movements are a result of human-object interation and simulation. At each time step, we use a per-step model ${\bs \pi}^{\text{step}}_{\text{KIN}}$ to compute the next frame pose based on the next frame observations: we obtain an egocentric input vector $\widetilde{\bs{c}}_t$ through the agent-centric transformation function $\widetilde{\bs{c}}_t = \bs T_{\text{AC}}(\widetilde{\bs{q}}_t, \widetilde{\bs o_{t+1}}, \widetilde{\bs h}_{t+1} /  \bs \phi_{t+1})$ where $\Tilde{\bs{c}}_t \triangleq ( \widetilde{\bs{r}}^{\prime rot}_{t},\widetilde{\bs{j}}^{\text{rot}}_t, \widetilde{\bs o^{\prime}}_{t+1}, \widetilde{\bs h}^{\prime}_{t+1}/\bs{\phi}_{t+1})$ contains the current agent-centric root orientation $\widetilde{\bs{r}}^{\prime rot}_{t}$, joint angles $\widetilde{\bs{j}}^{\text{rot}}_t$, and image feature for next frame $\bs{\phi}_{t+1}$, object state $\widetilde{\bs{o}}^{\prime}_{t+1}$ or camera pose $\widetilde{\bs{h}}^{\prime}_{t+1}$. From $\widetilde{\bs c}_{t}$, the kinematic policy ${\bs \pi}^{\text{step}}_{\text{KIN}}$ computes the root angular velocity $\widetilde{\bs w}_t$, linear velocity $\widetilde{\bs v}_t$, and next frame joint rotation $\widetilde{\bs j}^{\text{rot}}_{t+1}$. The next frame pose is computed through a finite integration module $\bs T_{\text{finite}}$ with time difference $\delta t  = 1/30s$:

\vspace{-2mm}
\begin{small}
\begin{align}
    \widetilde{\bs{\omega}}_t, \widetilde{\bs{v}}_t, \widetilde{\bs{j}}^{\text{rot}}_{t+1} =  \bs{\pi}_{\text{KIN}}^{\text{step}}(\widetilde{\bs{c}}_t), \quad & \widetilde{\bs q}_{t+1} = \bs T_{\text{finite}}(\widetilde{\bs{\omega}}_t, \widetilde{\bs{v}}_t, \widetilde{\bs{j}}^{\text{rot}}_{t+1}, \widetilde{\bs q}_t).
\end{align}
\end{small}
\vspace{-4mm}

    \begin{algorithm}[tb]
     \small
      \caption{Learning kinematic policy via supervised learning.} \label{alg:sl}
      \begin{algorithmic}[1]
       \State \textbf{Input:}  Egocentric videos $\bs I$ and paired ground truth motion dataset $ \bs{\widehat Q}$
      \While{not converged}{
      \State ${\bs M}_{\text{SL}}$ $\leftarrow \emptyset $ \Comment{initialize sampling memory} 
           \While{M not full}{
                \State $\bs I_{1:t} \leftarrow$ random sequence of images $\bs I_{1:T}$ from the dataset $\bs I$
                \State $\bs{\widetilde{q}}_1, \widetilde{\bs o}_{1},\widetilde{\bs h}_{1:T} / \bs{\phi}_{1:T} = \bs{\pi}_{\text{KIN}}^{\text{init}}({\bs I_{1:T}}) \quad $  \Comment{compute scene context and initial pose} 
                \For{$i \leftarrow 1...T$}
                    \State $\widetilde{\bs{c}}_t \leftarrow \bs T_{\text{AC}}({\widetilde{\bs{q}}}_t, \widetilde{\bs o}_{t+1}, \widetilde{\bs h}_{t+1} / \bs \phi_{t+1})$ \Comment{compute agent-centric input features} 
                    \State $\widetilde{\bs{q}}_{t+1} \leftarrow \bs T_{\text{finite}}(\bs{\pi}_{\text{KIN}}^{\text{step}}(\widetilde{\bs{c}}_t), \widetilde{\bs{q}}_t)$ 
                    \State store $(\bs{\widetilde{q}}_{t}, \bs{\widehat{q}}_{t})$ into memory ${\bs M}_{\text{SL}}$
                \EndFor
            }
            \EndWhile
        } 
        \State $\bs{\pi}_{\text{KIN}}^{\text{step}}, \bs{\pi}_{\text{KIN}}^{\text{init}} \leftarrow$ supervised learning update using data collected in ${\bs M}_{\text{SL}}$ for 10 epoches. 
        \EndWhile
      \end{algorithmic}
\end{algorithm}
When trained without physics simulation, we auto-regressively apply the kinematic policy and use the computed $\widetilde{\bs q}_{t+1} $ as the input for the next timestep. This procedure is outlined at Alg. \ref{alg:sl}. Since all mentioned calculations are end-to-end differentiable, we can  directly optimize our ${\bs \pi}^{\text{init}}_{\text{KIN}}$ and ${\bs \pi}^{\text{step}}_{\text{KIN}}$ through supervised learning. Specifically, given ground truth $\widehat{\bs{q}}_{1:T}$ and estimated $\widetilde{\bs{q}}_{1:T}$ pose sequence, our loss is computed as the difference between the desired and ground truth values of the following quantities: agent root position ($\widehat{\bs{{r}}}^{\text{pos}}_{t}$ vs $\widetilde{\bs{{r}}}^{\text{pos}}_{t}$) and orientation ($\widehat{\bs{{r}}}^{\text{rot}}_{t}$ vs $\widetilde{\bs{{r}}}^{\text{rot}}_{t}$), agent-centric object position ($\widehat{\bs{o}}^{\prime \text{pos}}_{t}$ vs $\widetilde{\bs{o}}^{\prime \text{pos}}_{t}$) and orientation  ($\widehat{\bs{o}}^{\prime \text{rot}}_{t}$ vs $\widetilde{\bs{o}}^{\prime \text{rot}}_{t}$), and agent joint orientation ($\widehat{\bs{j}}^{\text{rot}}_{t}$ vs $\widetilde{\bs{j}}^{\text{rot}}_{t}$) and position ($\widehat{\bs{j}}^{\text{pos}}_{t}$ vs $\widetilde{\bs{j}}^{\text{pos}}_{t}$, computed using forward kinematics): 

\vspace{-4mm}
\begin{small}
\begin{equation}
\mathcal{L}_{\text{SL}} = \sum_{i=1}^{T}\| \widetilde{\bs{r}}^{\text{rot}}_{t} \ominus \widehat{\bs{{r}}}^{\text{rot}}_{t} \|^{2} + \| \widetilde{\bs{r}}^{\text{pos}}_{t}{-}\widehat{\bs{{r}}}^{\text{pos}}_{t} \|^{2} + \|\widetilde{\bs{o}}^{\prime \text{rot}}_{t} \ominus \widehat{\bs{{o}}}^{\prime \text{rot}}_{t}  \|^{2} 
+ \|  \widetilde{\bs{o}}^{\prime \text{pos}}_{t}{-}\widehat{\bs{{o}}}^{\prime \text{pos}}_{t} \|^{2} +
\| \widetilde{\bs{j}}^{\text{rot}}_{t}{\ominus}\widehat{\bs{j}}^{\text{rot}}_{t} \|^{2} + \| \widetilde{\bs{j}}^{\text{pos}}_{t}{-}\widehat{\bs{j}}^{\text{pos}}_{t} \|^{2}.      
\label{equ:loss_bc}
\end{equation}
\end{small}

\vspace{-3mm}
\subsection{Dynamics-Regulated Training}
\label{sec:dynamics_regulated}
\vspace{-2mm}

To tightly integrate our kinematic and dynamics models, we design a \emph{dynamics-regulated training} procedure, where the kinematic policy learns from explicit physics simulation. In the procedure described in the previous section, the next-frame pose fed into the network is computed through finite integration and is not checked by physical laws: whether a real human can perform the computed pose is never verified. Intuitively, this amounts to mentally think about moving in a physical space \emph{without actually moving}. Combining our UHC and our kinematic policy, we can leverage the prelearned motor skills from UHC and let the kinematic policy act directly in a simulated physical space to obtain feedback about physical plausibility. The procedure for dynamics-regulated training is outlined in Alg. \ref{alg:dynamics_regulated}. In each episode, we use  $\bs{\pi}^{\text{init}}_{\text{KIN}}$ and $\bs{\pi}^{\text{step}}_{\text{KIN}}$ as in Alg. \ref{alg:sl}, with the key distinction being: at the next timestep $t+1$, the input to the kinematic policy is the result of UHC and physics simulation $\bs{q}_{t+1} $ instead of $\widetilde{\bs{q}}_{t+1}$. $\bs{q}_{t+1}$ explicitly verify that the $\widetilde{\bs{q}}_{t+1}$ produced by the kinematic policy can be successfully followed by a motion controller. Using $\bs{q}_{t+1}$ also informs our $\bs{\pi}^{\text{step}}_{\text{KIN}}$ of the current humanoid state and encourages the policy to adjust its predictions to improve humanoid stability.

\begin{algorithm}[tb]
\footnotesize
  \caption{Learning kinematic policy via dynamics-regulated training.} \label{alg:dynamics_regulated}
  \begin{algorithmic}[1]
   \State \textbf{Input:}  Pre-trained controller ${\bs \pi}_{\text{UHC}}$, egocentric videos $\bs I$, and paired ground truth motion dataset $ \bs{\widehat Q}$ 
  \State Train $\bs{\pi}_{\text{KIN}}^{\text{init}}$, $\bs{\pi}_{\text{KIN}}^{\text{step}}$ using Alg. \ref{alg:sl} for 20 epoches (optional). 
  \While{not converged}{
  \State ${\bs M}_{\text{dyna}}$ $\leftarrow \emptyset $ \Comment{initialize sampling memory} 
       \While{${\bs M}_{\text{dyna}}$ not full}{
          \State $\bs I_{1:t} \leftarrow$ random sequence of images $\bs I_{1:T}$
            \State $\bs{{q}}_1 \leftarrow \bs{\widetilde{q}}_1 , \widetilde{\bs o}_{1},\widetilde{\bs h}_{1:T} / \bs{\phi}_{1:T}\leftarrow \bs{\pi}_{\text{KIN}}^{\text{init}}({\bs I_{1:T}}) \quad $  \Comment{compute scene context and initial pose} 
            \State $\bs s_1 \leftarrow \left({\bs{q}}_{1}, \dot{\bs{q}}_{1} \right)$\Comment{compute intial state for simulation}
            \For{$i \leftarrow 1...T$}
                \State ${\bs{c}}_t \leftarrow \bs T_{\text{AC}}({\bs{q}}_t, \widetilde{\bs o}_{t+1}, \widetilde{\bs h}_{t+1}/\bs \phi_{t+1})$ \Comment{compute agent-centric features using simulated pose \textbf{${\bs{q}}_t$}} 
                \State $\widetilde{\bs{q}}_{t+1} \sim \bs T_{\text{finite}}(\bs{\pi}_{\text{KIN}}^{\text{step}}({\bs{c}}_t), {\bs{q}}_t)$  \Comment{sample from $\bs{\pi}_{\text{KIN}}^{\text{step}}$ as a guassian policy} 
                \State $\bs s_t \leftarrow \left({\bs{q}}_{t}, \dot{\bs{q}}_{t} \right)$ 
                \State $\bs s_{t+1} \leftarrow {P}_{\text{physics}}(\bs s_{t+1} |\bs s_{t}, \bs a_t)$, $\bs a_t \leftarrow {\bs \pi}_{\text{UHC}}(\bs a_t | \bs s_t, \widetilde{\bs{q}}_{t+1})$  \Comment{phyics simulation using  ${\bs \pi}_{\text{UHC}}$ }
                \State  $\bs q_{t+1} \leftarrow \bs s_{t+1}$, $\bs r^{\text{KIN}}_t \leftarrow $ reward from Eq. \ref{eq:ft_reward} \Comment{extract reward and $\bs q_{t+1}$ from simulation}
                \State store $(\bs s_{t}, \bs a_t, \bs r_t, \bs s_{t+1}, \bs{\widehat{q}}_{t}, \widetilde{\bs{q}}_{t+1})$ into memory ${\bs M}_{\text{dyna}}$
            \EndFor
        }
        \EndWhile
    } 
    \State $\bs{\pi}_{\text{KIN}}^{\text{step}} \leftarrow$ Reinforcement learning updates using experiences collected  in  ${\bs M}_{\text{dyna}}$ for 10 epoches. 
    \State $\bs{\pi}_{\text{KIN}}^{\text{init}}$, $\bs{\pi}_{\text{KIN}}^{\text{step}} \leftarrow$ Supervised learning update using experiences collected in ${\bs M}_{\text{dyna}}$ for 10 epoches. 
    \EndWhile
  \end{algorithmic}
\end{algorithm}

\noindent \textbf{Dynamics-regulated optimization.} Since the physics simulation is not differentiable, we cannot directly optimize the simulated pose $\bs q_t$; however, we can optimize $\bs q_t$ through reinforcement learning and $\widetilde{\bs q}_t$ through supervised learning. Since we know that $\widehat{\bs q_t}$ is a \emph{good guess} reference motion for UHC, we can directly optimize $\widetilde{\bs q}_t$ via supervised learning as done in Sec. \ref{sec:kin_policy} using the loss defined in Eq. \ref{equ:loss_bc}. Since the data samples are collected through physics simulation, the input $\bs q_t$ is  physically-plausible and more diverse than those collected purely through auto-regressively applying $\bs{\pi}_{\text{KIN}}^{\text{step}}$ in Alg. \ref{alg:sl}. This way, our dynamics-regulated training procedure performs a powerful data augmentation step, exposing $\bs{\pi}_{\text{KIN}}^{\text{step}}$ with diverse states collected from simulation.

 However, MoCap pose $\widehat{\bs q}_{t}$ is imperfect and can contain physical violations itself (foot-skating, penetration \etc), so asking the policy $\bs{\pi}_{\text{KIN}}^{\text{step}}$ to produce $\widehat{\bs q}_{t}$ as reference motion \emph{regardless of the current humanoid state} can lead to instability and cause the humanoid to fall. The kinematic policy should adapt to the current simulation state and provide reference motion $\widetilde{\bs q}_{t}$ that can lead to poses similar to $\widehat{\bs q}_{t}$ yet still physically-plausible. Such behavior will not emerge through supervised learning and require \emph{trial and error}. Thus, we optimize $\bs{\pi}_{\text{KIN}}^{\text{step}}$ through reinforcement learning and reward maximization. We design our RL reward to have two components: motion imitation and dynamics self-supervision. The motion imitation reward encourages the policy to match the computed camera trajectory $\widetilde{\bs h}_{t}$ and MoCap pose $\widehat{\bs q}_{t}$, and serves as a regularization on motion imitation quality. The dynamics self-supervision reward is based on the insight that the  disagreement between $\widetilde{\bs{q}}_{t}$ and ${\bs{q}}_{t}$ contains important information about the quality and physical plausibility of $\widetilde{\bs{q}}_{t}$: the better $\widetilde{\bs{q}}_{t}$ is, the easier it should be for UHC to mimic it. Formally, we define the reward for $\bs{\pi}_{\text{KIN}}^{\text{step}}$  as:
\begin{equation}
\begin{aligned}
\label{eq:ft_reward}
{\bs r}_t &=  w_{\text{hp}}  e^{-45.0( \|{\bs{h}^{\text{pos}}}_t - \widetilde{\bs{h}}^{\text{pos}}_t \|^2)}
+  w_{\text{hq}} e^{-45.0( \|{\bs{h}^{\text{rot}}}_t \ominus \widetilde{\bs{h}}^{\text{rot}}_t \|^2)} + w^{\text{gt}}_{\text{jv}} e^{ -0.005(\|\bs{\dot{j}}^{\text{rot}}_{t}\ominus\widehat{\bs{\dot{j}}^{\text{rot}}_{t}}\|^{2})}+  \\ 
& w_{\text{jr}}^{\text{gt}}  e^{-50.0( \|{\bs{j}}^{\text{rot}}_t \ominus \widehat{\bs{j}}^{\text{rot}}_t \|^2)} + w_{\text{jr}}^{\text{dyna}} e^{-50.0( \|{\bs{j}^{\text{rot}}} \ominus \widetilde{\bs{j}}^{\text{rot}}_{t} \|^2)} +
 w^{\text{dyna}}_{\text{jp}} e^{-50.0(\|\bs{j}^{\text{pos}}_{t}-\widetilde{\bs{j}}^{\text{pos}}_{t}\|^{2})}, 
\end{aligned}
\end{equation}
\vspace{-3mm}

$ w_{\text{hp}}$, $  w_{\text{hq}}$ are weights for matching the extracted camera position $ \widetilde{\bs h}^{\text{pos}}_t$ and orientation $\widetilde{\bs h}^{\text{rot}}_t$; $w^{\text{gt}}_{jr}, w^{\text{gt}}_{\text{jv}}$ are for matching ground truth joint angles ${ \widehat{\bs j}}^{\text{rot}}$ and angular velocities $\widehat{\bs{\dot{j}}^{\text{rot}}_{t}}$. $ w_{\text{jr}}^{\text{dyna}}$, $ w_{\text{jp}}^{\text{dyna}}$ are weights for the dynamics self-supervision rewards, encouraging the policy to match the target kinematic joint angles $\widetilde{\bs{j}}^{\text{rot}}_{t}$ and positions $\widetilde{\bs{j}}^{\text{pos}}_{t}$ to the simulated joint angles ${\bs{j}}^{\text{rot}}_{t}$ and positions ${\bs{j}}^{\text{pos}}_{t}$. As demonstrated in Sec. \ref{sec:abla}, the RL loss is particularly helpful in adapting to challenging real-world sequences, which requires the model to adjust to domain shifts and unseen motion.

\noindent \textbf{Test-time.} At the test time, we follow the same procedure outlined in Alg \ref{alg:dynamics_regulated} and Fig.\ref{fig:overview} to roll out our policy to obtain simulated pose ${\bs{q}}_{1:T}$ given a sequence of images $\bs I_{1:T}$. The difference being instead of sampling from $\bs{\pi}_{\text{KIN}}^{\text{step}} (\bs c_t)$ as a Guassian policy, we use the mean action directly.

 \begin{figure*}[tb]
\begin{center}
\includegraphics[width=\textwidth]{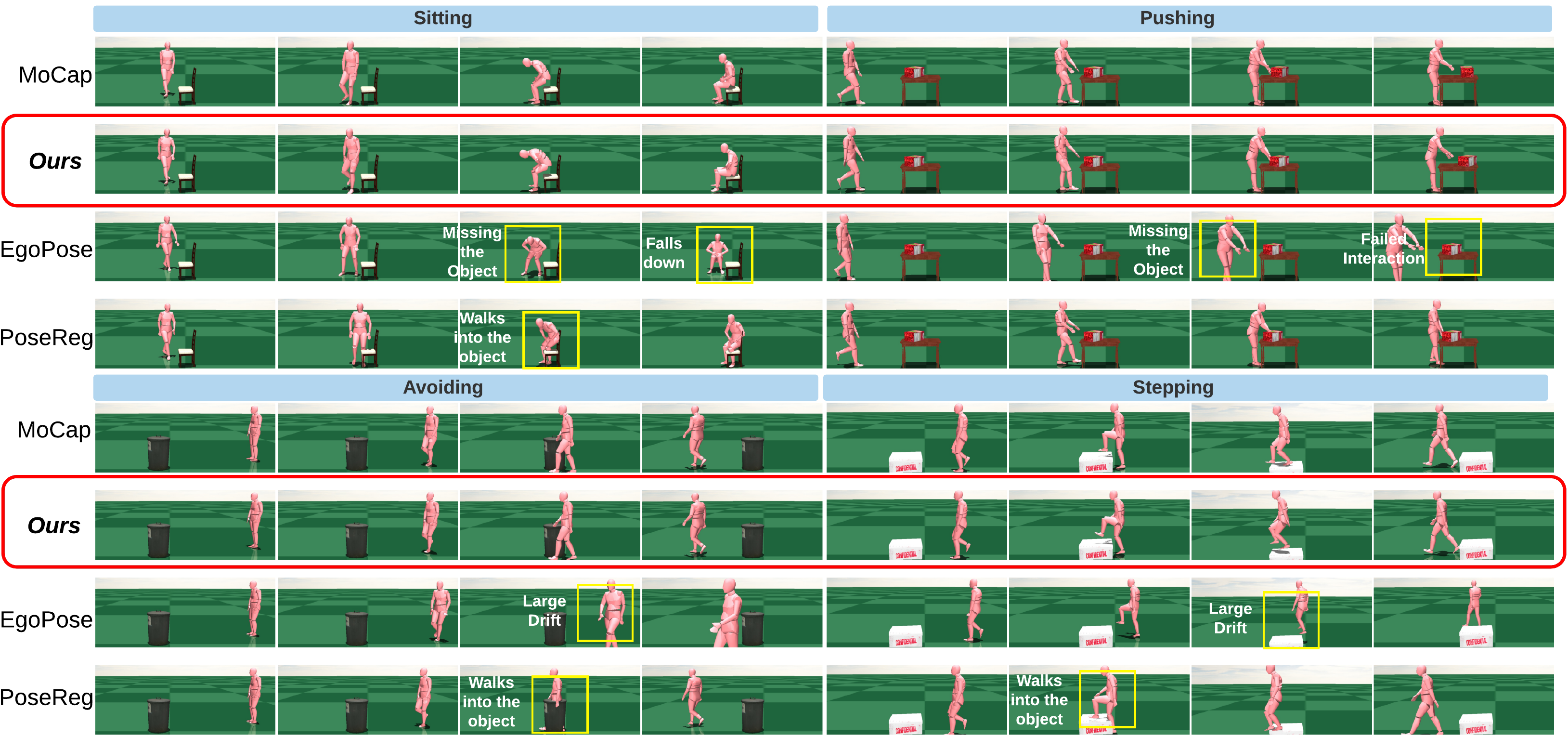}
\end{center}
\vspace{-2mm}
\caption{Results of egocentric pose and human-object interaction estimation from the MoCap datset.}
\label{fig:qualitative}
\vspace{-3mm}
\end{figure*}

\vspace{-3mm}
\section{Experiments}
\vspace{-2mm}

\noindent \textbf{Datasets.} As no public dataset contains synchronized ground-truth full-body pose, object pose, and egocentric videos with human-object interactions, we record two egocentric datasets: one inside a MoCap studio, another in the real-world. The MoCap dataset contains 266 sequences (148k frames) of paired egocentric videos and annotated poses. It features one of the five actions: sitting down on a chair, avoiding obstacles, stepping on a box, pushing a box, and generic locomotion \cite{Yuan_2018_ECCV} (walking, running, crouching) recorded using a head-mounted GoPro. Each action has around $50$ sequences with different starting position and facing, gait, speed \etc. We use an $80\mbox{--}20$ train test data split on this MoCap dataset. The real-world dataset is only for testing purpose and contains 183 sequences (50k frames) of an additional subject performing similar actions in an everyday setting wearing a head-mounted iPhone. For both datasets, we use different objects and varies the object 6DoF pose for each capture take. Additional details (diversity, setup \etc) can be found in Appendix \ref{app:data}.

\noindent \textbf{Evaluation metrics.} We use both {pose-based} and {physics-based} metrics for evaluation. To evaluate the 3D global pose accuracy, we report the root pose error ($ \text{E}_{\text{root}}$) and root-relative mean per joint position error \cite{Kocabas2020VIBEVI} ($\text{E}_{\text{mpjpe}}$). When ground-truth root/pose information is unavailable (for real-world dataset), we substitute $ \text{E}_{\text{root}}$ with $\text{E}_{\text{cam}}$ to report camera pose tracking error. We also employ four physics based pose metrics: acceleration error ($\text{E}_{\text{acc}}$), foot skating ($\text{ FS}$), penetration ($\text{PT}$), and interaction success rate ($\text{S}_{\text{inter}}$). $\text{E}_{\text{acc}}$ (mm/frame$^2$) compares the ground truth and estimated average joint acceleration; $\text{FS}$ (mm) is defined the same as in Ling \etal \cite{Ling2020CharacterCU}; $\text{PT}$ (mm) measures the average penetration distance between our humanoid and the scene (ground floor and objects). Notice that our MoCap dataset has \emph{an penetration of 7.182 mm and foot sliding of 2.035 mm} per frame, demonstrating that the MoCap data is imperfect and may not serve as the best target motion. $\text{S}_{\text{inter}}$ is defined as whether the objects of interest has been moved enough (pushing and avoiding) or if desired motion is completed (stepping and sitting). If the humanoid falls down at any point, $\text{S}_{\text{inter}} = 0$.  For a full definition of our evaluation metrics, please refer to Appendix \ref{app:kin_poly}.

\noindent \textbf{Baseline methods.} To show the effectiveness of our framework, we compare with the previous state-of-the-art  egocentric pose estimation methods: (1) the best dynamics-based approach \emph{EgoPose} \cite{yuan2019ego} and (2) the best kinematics-based approach \emph{PoseReg}, also proposed in \cite{yuan2019ego}. We use the official implementation and augment their input with additional information ($\widetilde{\bs o_t}$ or $\widetilde{\bs h}_t$) for a fair comparison. In addition, we incorporate the fail-safe mechanism \cite{yuan2019ego} to reset the simulation when the humanoid loses balance to ensure the completion of each sequence (details in Appendix \ref{app:kin_poly}).

\textbf{Implementation details.} We use the MuJoCo free physics simulator \cite{Todorov2012MuJoCo} and run the simulation at 450 Hz. Our learned policy is run every 15 timesteps and assumes that all visual inputs are at 30 Hz. The humanoid follows the kinematic and mesh definition of the SMPL model and has 25 bones and 76 DoF. We train our method and baselines on the training split (202 sequences) of our MoCap dataset. The training process takes about 1 day on an RTX 2080-Ti with 35 CPU threads. After training and the initialization step, our network is causal and runs at 50 FPS on an Intel desktop CPU. The main evaluation is conducted using head poses, and we show results on using the image features in ablation. For more details on the implementation, refer to Appendix \ref{app:kin_poly} and \ref{app:uhc}.

\vspace{-3mm}
\subsection{Results}
\vspace{-2mm}

\textbf{MoCap dataset results.} Table \ref{tab:quantativeMoCapWild} shows the quantitative comparison of our method with the baselines. All results are averaged across five actions, and all models have access to the same inputs. We observe that our method, trained either with supervised learning or dynamics-regulated, outperform the two state-of-the-art methods across all metrics. Not surprisingly, our purely kinematic model performs the best on pose-based metrics, while our dynamics-regulated trained policy excels at the physics-based metrics. Comparing the kinematics-only models we can see that our method has a much lower (79.4\% error reduction) root and joint position error (62.1\% error reduction) than PoseReg, which shows that our object-aware and autoregressive design of the kinematic model can better utilize the provided visual and scene context and avoid compounding errors. Comparing with the dynamics-based methods, we find that the humanoid controlled by EgoPose has a much larger root drift, often falls down to the ground, and has a much lower success rate in human-object interaction (48.4 \% vs 96.9\%). Upon visual inspection in Fig. \ref{fig:qualitative}, we can see that our kinematic policy can faithfully produce human-object interaction on almost every test sequence from our MoCap dataset, while PoseReg and EgoPose often miss the object-of-interest (as can be reflected by the large root tracking error). Both of the dynamics-based methods has smaller acceleration error, foot skating, and penetration; some even smaller than MoCap (which has 2 mm $\text{FS}$ and 7mm $\text{PT}$). Notice that our joint position error is relatively low compared to state-of-the-art third-person pose estimation methods \cite{Kocabas2020VIBEVI, Kolotouros2019LearningTR,Luo20203DHM} due to our strong assumption about known object of interest, its class, and potential human-object interactions, which constrains the ill-posed problem pose estimation from just front-facing cameras.

\begin{table}[t]
\caption{Quantitative results on pose and physics based metrics on the MoCap and real-world Dataset.} \label{tab:quantativeMoCapWild}
\centering
\resizebox{\linewidth}{!}{%
\begin{tabular}{lrrrrrrr}
\toprule
\multicolumn{8}{c}{MoCap dataset } 
\\ 
\midrule
Method  & Physics& $\text{S}_{\text{inter}} \uparrow$ & $\text{E}_{\text{root}}\downarrow$ & $\text{E}_{\text{mpjpe}} \downarrow$   & $\text{E}_{\text{acc}} \downarrow$  & $\text{FS} \downarrow$ & $\text{PT} \downarrow$    \\ \midrule
PoseReg       & \xmark &{-}& {0.857} &{87.680} & {12.981} & {8.566} & {42.153}  \\ 
\text{Kin\_poly: supervised learning (ours)} & \xmark &{-} & \textbf{0.176} & \textbf{33.149}  & \textbf{6.257} & {5.579}&{10.076} \\    \midrule 
EgoPose       & \cmark  & {48.4\%} & {1.957} & {139.312}  & {9.933} & \textbf{2.566} & {7.102} \\ 
\text{Kin\_poly: dynamics-regulated (ours)}   & \cmark & \textbf{96.9\%}  & {0.204} & {39.575} & {6.390} & {3.075} &  \textbf{0.679} 
\\ 
\bottomrule 
\end{tabular}}\\ 

 \centering    
\resizebox{\linewidth}{!}{%
    \begin{tabular}{lrrrrr|cccc}
    \toprule 
    \multicolumn{10}{c}{Real-world dataset} 
    \\ \midrule
    Method  & Physics & $\text{S}_{\text{inter}} \uparrow$ & $\text{E}_{\text{cam}}\downarrow$  & $\text{FS} \downarrow$ & $\text{PT} \downarrow$  & \multicolumn{4}{|c}{Per class success rate $\text{S}_{\text{inter}} \uparrow$}  \\ \midrule
    PoseReg       & \xmark & {-}  & {1.260} &  {6.181} & {50.414} & \multirow{2}{*}{Sit} & \multirow{2}{*}{Push} & \multirow{2}{*}{Avoid} & \multirow{2}{*}{Step}\\ 
    \text{Kin\_poly: supervised learning (ours)}   & \xmark & {-} & {0.491} & {5.051} &  {34.930} \\
    \cmidrule(lr){1-6} \cmidrule(lr){7-10}
    EgoPose       & \cmark & {9.3}\%  & {1.896}  & \textbf{2.700} & {1.922} &{7.93}\%  & {6.81}\%&{4.87}\% &{0.2}\% \\
    \text{Kin\_poly: dynamics-regulated (ours)}   & \cmark & \textbf{93.4\%} & \textbf{0.475}  & {2.726} &  \textbf{1.234}  & \textbf{98.4}\%  & \textbf{95.4}\%& \textbf{ 100}\% & \textbf{74.2}\%  \\ 
    \bottomrule 
    \end{tabular}}
    
\vspace{-3mm}
\end{table}

\textbf{Real-world dataset results.} The real-world dataset is far more challenging, having similar number of sequences (183 clips) as our training set (202 clips) and recorded using different equipment, environments, and motion patterns. Since no ground-truth 3D poses are available, we report our results on camera tracking and physics-based metrics. As shown in Table \ref{tab:quantativeMoCapWild}, our method outperforms the baseline methods by a large margin in almost all metrics: although EgoPose has less foot-skating  (as it also utilizes a physics simulator), its human-object interaction success rate is extremely low. This can be also be reflected by the large camera trajectory error, indicating that the humanoid is drifting far away from the objects. The large drift can be attributed to the domain shift and challenging locomotion from the real-world dataset, causing EgoPose's humanoid controller to accumulate error and lose balance easily. On the other hand, our method is able to generalize and perform successful human-object interactions, benefiting from our pretrained UHC and kinematic policy's ability to adapt to new domains and motion. Table \ref{tab:quantativeMoCapWild} also shows a success rate breakdown by action. Here we can see that ``stepping on a box" is the most challenging action as it requires the humanoid lifting its feet at a precise moment and pushing itself up. Note that our UHC has never been trained on any stepping or human-object interaction actions (as AMASS has no annotated object pose) but is able to perform these action. As motion is best seen in videos, we refer readers to our \href{https://zhengyiluo.github.io/projects/kin\_poly/}{supplementary video}.

\begin{wraptable}{r}{0.5\linewidth}
\vspace{-3mm}
\caption{Ablation study of different components of our framework.}
\raggedleft
\resizebox{\linewidth}{!}{%
\begin{tabular}{lccccccc}
\toprule
\multicolumn{4}{c}{Component} & \multicolumn{4}{c}{Metric} \\ 
\cmidrule(lr){1-4} \cmidrule{5-8} 
 SL & Dyna\_reg &  RL & VIO & $\text{S}_{\text{inter}} \uparrow$  & $\text{E}_{\text{cam}}\downarrow$ & $\text{FS} \downarrow$ & $\text{PT} \downarrow$   \\ \midrule
\cmark    & \xmark & \xmark & \cmark& {73.2\%} & {0.611}  & {4.234} &  {1.986}  \\
\cmark    & \cmark & \xmark & \cmark & {80.9\%} & {0.566}   & {3.667} &  {4.490}  \\
\cmark    & \cmark & \cmark & \xmark & {54.1\%} & {1.129}  & {7.070} &  {5.346}  \\
\cmark    & \cmark & \cmark & \cmark & \textbf{93.4\%} & \textbf{0.475}  & \textbf{2.726} &  \textbf{1.234}  \\
\bottomrule 
\end{tabular}
}
\label{tab:ablation}
\vspace{-3mm}
\end{wraptable}

\vspace{-3mm}
\subsection{Ablation Study}
\vspace{-2mm}
\label{sec:abla}
To evaluate the importance of our components, we train our kinematic policy under different configurations and study its effects on the \emph{real-world dataset}, which is much harder than the MoCap dataset. The results are summarized in Table~\ref{tab:ablation}. Row 1 (R1) corresponds to training the kinematic policy only with Alg. \ref{alg:sl} only and use UHC to mimic the target kinematic motion as a post-processing step. Row 2 (R2) are the results of using dynamics-regulated training but only performs the supervised learning part. R3 show a model trained with optical flow image features rather than the estimated camera pose from VIO. Comparing R1 and R2, the lower interaction success rate (73.2\% vs 80.9\%) indicates that exposing the kinematic policy to states from the physics simulation serves as a powerful data augmentation step and leads to a model more robust to real-world scenarios. R2 and R4 show the benefit of the RL loss in dynamics-regulated training: allowing the kinematic policy to deviate from the MoCap poses makes the model more adaptive and achieves higher success rate. R3 and R4 demonstrate the importance of \emph{intelligently} incorporating extracted camera pose as input: visual features $\bs \phi_t$ can be noisy and suffer from domain shifts, and using techniques such as SLAM and VIO to extract camera poses as an additional input modality can largely reduce the root drift. Intuitively, the image features computed from optical flow and the camera pose extracted using VIO provide a similar set of information, while VIO provides a cleaner information extraction process. Note that our kinematic policy \emph{without using} extracted camera trajectory outperforms EgoPose that \emph{uses camera pose} in both success rate and camera trajectory tracking. Upon visual inspection, the humanoid in R3 largely does not fall down (compared to EgoPose) and mainly attributes the failure cases to drifting too far from the object.

\section{Discussions}
\vspace{-1mm}
\label{app:discuss}
\subsection{Failure Cases and Limitations} 
\vspace{-1mm}
Although our method can produce realistic human pose and human-object interaction estimation from egocentric videos, we are still at the early stage of this challenging task. Our method performs well in the MoCap studio setting and generalizes to real-world settings, but is limited to a \emph{predefined set} of interactions where we have data to learn from. Object class and pose information is computed by off-the-shelf methods such as Apple's ARkit \cite{ARKitPose}, and is provided as a strong prior to our kinematic policy to infer pose. We also only factor in the 6DoF object pose in our state representation and discard all other object geometric information. The lower success rate on the real-world dataset also indicates that our method still suffers from covariate shifts and can become unstable when the shift becomes too extreme. Our Universal Humanoid Controller can imitate everyday motion with high accuracy, but can still fail at extreme motion. Due to the challenging nature of this task, in this work, we focus on developing a general framework to ground pose estimation with physics by merging the kinematics and dynamics aspects of human motion. To enable pose and human-object interaction estimation for arbitrary actions and objects, better scene understanding and kinematic motion planning techniques need to be developed. 

\subsection{Conclusion and Future Work} 
\vspace{-1mm}
In this paper, we tackle, for the first time, estimating physically-plausible 3D poses from an egocentric video while the person is interacting with objects. We collect a motion capture dataset and real-world dataset to develop and evaluate our method, and extensive experiments have shown that our method outperforms all prior arts. We design a dynamics-regulated kinematic policy that can be directly trained and deployed inside a physics simulation, and we purpose a general-purpose humanoid controller that can be used in physics-based vision tasks easily. Through our real-world experiments, we show that it is possible to estimate 3D human poses and human-object interactions from just an egocentric view captured by consumer hardware (iPhone). In the future, we would like to support more action classes and further improve the robustness of our method by techniques such as using a learned motion prior. Applying our dynamics-regulated training procedure to other vision tasks such as visual navigation and third-person pose estimation can also be of interest.

\textbf{Acknowledgements:} This project was sponsored in part by IARPA  (D17PC00340), and JST AIP Acceleration Research Grant (JPMJCR20U1).

{\small
\bibliographystyle{plain}
\bibliography{main}
}

\newpage

\newpage

\appendix
{   
    \hypersetup{linkcolor=black}
    \begin{Large}
        \textbf{Appendix}
    \end{Large}
    \etocdepthtag.toc{mtappendix}
    \etocsettagdepth{mtchapter}{none}
    \etocsettagdepth{mtappendix}{subsection}
    \newlength\tocrulewidth
    \setlength{\tocrulewidth}{1.5pt}
    \parindent=0em
    \etocsettocstyle{\vskip0.5\baselineskip}{}
    \tableofcontents
}

\input{Appendix}

\end{document}

%% file: appendix.tex
\vspace{-3mm}
\section{Qualitative Results (Supplemantry Video)}
\vspace{-1mm}
\label{app:video}
As motion is best seen in videos, we provide extensive qualitative evaluations in the \href{https://zhengyiluo.github.io/projects/kin\_poly/}{supplementary video}. Here we list a timestamp reference for evaluations conducted in the video: 
\begin{itemize}
    \item Qualitative results from real-world videos (00:12).
    \item Comparison with the state-of-the-art methods on the MoCap dataset's test split (01:27).
    \item Comparison with the state-of-the-art methods on the real-world dataset (02:50).
    \item Failure cases for the dynamics-regulated kinematic policy (04:23).
    \item Qualitative results from the Universal Humanoid Controller (UHC) (4:39).
    \item Failure cases for the UHC (05:20).
\end{itemize}

\vspace{-3mm}
\section{Dynamics-regulated Kinematic Policy}
\vspace{-2mm}
\label{app:kin_poly}

\subsection{Evaluation Metrics Definition}
\vspace{-2mm}
Here we provide details about our proposed evaluation metrics: 
\begin{itemize}
    \item Root error: $\bf E_{\text{root}}$ compares the estimated and ground truth root rotation and orientation, measuring the difference in the respective $4 \times 4$ transformation matrix ($\bs M_{t}$): $\frac{1}{T} \sum_{t=1}^{T}\|I-(\bs M_{t} \widehat{\bs M}_{t}^{-1})\|_{F}$. This metric reflects both the position and orientation tracking quality. 
    \item Mean per joint position error: $ \bf E_{\text{mpjpe}}$ (mm) is the popular 3D human pose metric \cite{Kocabas2020VIBEVI, Kanazawa2018EndtoEndRO, Kolotouros2019LearningTR} and is defined as $\frac{1}{J}\| {\bs j}^{\text{pos}} - \widehat{\bs j}^{\text{pos}} \|_2$ for $J$ number of joints. This value is root-relative and is computed after setting the root translation to zero.
    \item Acceleration error: $\text{E}_{\text{acc}}$ (mm/frame$^2$) measures the difference between the ground truth and estimated joint position acceleration: $\frac{1}{J}\| \ddot{\bs j}^{\text{pos}} - \widehat{\ddot{\bs j}}^{\text{pos}} \|_2$.
    \item Foot sliding: $\text{FS}$  (mm) is computed similarly as in \cite{Ling2020CharacterCU}, \ie ${\text{FS}} =d\left(2-2^{h / H}\right)$ where $d$ is the foot displacement and $h$ is the foot height of two consecutive poses. We use a height threshold of $H=33$ mm, the same as in \cite{Ling2020CharacterCU}.
    \item Penetration: $\text{PT}$  (mm) is provided by the physics simulation. It measures the per-frame average penetration distance between our simulated humanoid and the scene (ground and objects). Notice that Mujoco uses a soft contact model where a larger penetration will result in a larger repulsion force, so a small amount of penetration is expected. 
    \item Camera trajectory error: $\bf E_{\text{cam}}$ is defined the same as the root error, and measures the camera trajectory tracking instead of the root. To extract the camera trajectory from the estimated pose $\bs q_t$, we use the head pose of the humanoid and apply a delta transformation based on the camera mount's vertical and horizontal displacement from the head. 
    \item Human-object interaction success rate: $\text{S}_{\text{inter}}$ measures whether the desired human-object interaction is successful. If the humanoid falls down at any point during the sequence, the sequence is deemed unsuccessful. The success rate is measured automatically by querying the position, contact, and simulation states of the objects and humanoid. For each action: \begin{itemize}
        \item Sitting down: successful if the humanoid's pelvis or the roots of both legs come in contact with the chair at any point in time. 
        \item Pushing a box: successful if the box is moved more than 10 cm during the sequence. 
        \item Stepping on a box: successful if the humanoid's root is raised at least 10 cm off the ground and either foot of the humanoid has come in contact with the box. 
        \item Avoiding an obstacle: successful if the humanoid has not come in contact with the obstacle and the ending position of the root/camera is less than 50 cm away from the desired position  (to make sure the humanoid does not drift far away from the obstacle). 
    \end{itemize}
\end{itemize}

\subsection{Fail-safe during evaluation}
For methods that involve dynamics, the humanoid may fall down mid-episode and not complete the full sequence. In order to compare all methods fairly, we incorporate the ``fail-safe'' mechanism proposed in EgoPose \cite{yuan2019ego} and use the estimated kinematic pose to restart the simulation at the timestep of failure. Concretely, we measure point of failure by thresholding the difference between the reference joint position $\bs{\tilde j^{\text{pos}}_t}$ and the simulated joint position $\bs{ j^{\text{pos}}_t}$. To reset the simulation, we use the estimated kinematic pose $\bs{\tilde q_t}$ to set the simulation state.

\vspace{-2mm}
\subsection{Implementation Details} 
\vspace{-1mm}

The kinematic policy is implemented as a Gated Recurrent Unit (GRU) \cite{Cho2014LearningPR} based network with 1024 hidden units, followed by a three-layer MLP (1024, 512, 256) with ReLU activation. The value function for training the kinematic policy through reinforcement learning is a two-layer MLP (512, 256) with ReLU activation. We use a fixed diagonal covariance matrix and train for 1000 epoches using the Adam \cite{kingma2015Adam} optimizer. Hyperparameters for training  can be found in Table. \ref{tab:hyper_kin}:

\begin{table}[h]
\caption{Hyperparameters used for training the kinematic policy.} \label{tab:hyper_kin}
\centering
\resizebox{\linewidth}{!}{%
\begin{tabular}{lcccccccc}
\toprule
  & $\gamma $& Batch Size & Value Learning Rate & Policy Learning Rate  &  PPO clip $\epsilon$ & Covariance Std
  \\ \midrule
Value   & 0.95   & 10000 & $3\times 10^{-4}$ & $5\times 10^{-4}$ & 0.2 & $0.04$
\\ \midrule
& $ w_{\text{hp}}$ & $ w_{\text{hq}}$  & $ w^{\text{gt}}_{\text{jr}}$ & $ w^{\text{gt}}_{\text{jv}}$ & $ w^{\text{dyna}}_{\text{jr}}$  & $ w^{\text{dyna}}_{\text{jp}}$ 
  \\ \midrule
Value   & 0.15    & 0.15    & 0.2    & 0.1    & 0.2    & 0.2 \\
\bottomrule 
\end{tabular}}\\ 
\end{table}

\vspace{-2mm}
\subsection{Additional Experiments about Stochasticity} 
\vspace{-1mm}
Our kinematic policy is trained through physics simulation and samples a random sequence from the MoCap dataset for each episode. Here we study the stochasticity that rises from this process. We train our full pipeline with three different random seeds and report its results with error bars on both the MoCap test split and the real-world dataset. As can be seen in Table \ref{tab:stochastic}, our method has very small stochasticity and maintains high performance on \emph{both} the MoCap test split and the real-world dataset, demonstrating the robustness of our dynamics-regulated kinematic policy. Across different random seeds, we can see that ``stepping" is consistently the hardest action and ``avoiding" is the easiest. Intuitively, ``stepping" requires precise coordination between the kinematic policy and the UHC for lifting the feet and pushing up, while ``avoiding" only requires basic locomotion skills.

\begin{table}[t]
\caption{Results of our dynamics-regulated kinematic policy on the test split of MoCap and real-world datasets using different random seeds. The ``loco" motion in the MoCap dataset corresponds to the generic locomotion action, containing all sequences from the EgoPose \cite{yuan2019ego} Dataset. } \label{tab:stochastic}
\centering
\resizebox{\linewidth}{!}{%
    \begin{tabular}{cccccc|ccccc}
    \toprule 
    \multicolumn{11}{c}{MoCap dataset} 
    \\ \midrule
      \multirow{2}{*}{$\text{S}_{\text{inter}} \uparrow$} & \multirow{2}{*}{$\text{E}_{\text{root}}\downarrow$} & \multirow{2}{*}{$\text{E}_{\text{mpjpe}}\downarrow$} & \multirow{2}{*}{$\text{E}_{\text{acc}}\downarrow$}& \multirow{2}{*}{$\text{FS}  \downarrow$} & \multirow{2}{*}{$\text{PT}  \downarrow$} &   \multicolumn{5}{c}{Per class success rate $\text{S}_{\text{inter}} \uparrow$ }  \\  \cmidrule{7-11}
     & & &   &  & & Sit & Push & Avoid & Step & Loco \\ \midrule 
     {$ \footnotesize 96.87\% \pm 1.27\%$}& {$\footnotesize  0.21 \pm 0.01$} & {$ \footnotesize 39.46 \pm 0.52$} & {$ \footnotesize 6.27 \pm 0.1$} & {$\footnotesize 3.22 \pm 0.11 $}  & {$ \footnotesize 0.69 \pm 0.03 $} & {$\footnotesize  100\% $} &{$\footnotesize  97.20\% \pm 3.96\%$} & {$\footnotesize  100\% $} & {$ \footnotesize 86.70\% \pm 4.71\% $} & {$\footnotesize 97.4\% \pm 3.63\% $}\\ 
    \bottomrule 
\end{tabular}}

\resizebox{\linewidth}{!}{%
    \begin{tabular}{cccc|cccc}
    \toprule 
    \multicolumn{8}{c}{Real-world dataset} 
    \\ \midrule
     \multirow{2}{*}{$\text{S}_{\text{inter}} \uparrow$} & \multirow{2}{*}{$\text{E}_{\text{root}}\downarrow$} & \multirow{2}{*}{$\text{FS} \downarrow$} & \multirow{2}{*}{$\text{PT} \downarrow$} & \multicolumn{4}{c}{Per class success rate $\text{S}_{\text{inter}} \uparrow$ }  \\ \cmidrule{5-8}
      &  & & & Sit & Push & Avoid & Step  \\ \midrule 
      {$92.17\% \pm 1.41\%  $} & {$ 0.49 \pm 0.01$} &{$2.72 \pm 0.03 $} & {$1.03 \pm 0.16 $} & {$94.7\% \pm 4.20\% $} & {$93.10\% \pm  1.84\%$} & {$100.0\% $} & {$77.1 \% \pm 2.37\% $}  \\ 
     
    \bottomrule 
    \end{tabular}}

\vspace{-3mm}
\end{table}

\subsection{Additional Analysis into low Per Joint Error} 
As discussed in the results section, we notice that our Mean Per Joint Position Error is relatively low compared to third-person pose estimation methods, although egocentric pose estimation is arguably a more ill-posed task. To provide an additional analysis of this observation, here we report the per-joint positional errors for the four joints with the smallest and largest errors, in ascending order:

\begin{table}[h]
\caption{Per-joint error on the MoCap dataset} \label{tab:hyper_kin}
\centering
\resizebox{\linewidth}{!}{%
\begin{tabular}{lcccccccc}
\toprule
  Torso &	Left\_hip	&Right\_hip	&Spine	&Left\_toe	&Right\_toe	&Right\_hand	&Left\_hand
  \\ \midrule
7.099	& 8.064	& 8.380	 &15.167&65.060	&66.765&	74.599&	77.669
 \\
\bottomrule 
\end{tabular}}\\ 
\end{table}

As can be seen in the results, the toes and hands have much larger errors. This is expected as inferring hand and toe movements from only the egocentric view is challenging, and our network is able to extrapolate their position based on physical laws and prior knowledge of the scene context. Different from a third-person pose estimation setting, correctly estimating the torso area can be much easier from an egocentric point of view since torso movement is highly correlated with head motion. In summary, the low MPJPE reported on our MoCap dataset is the result of 1) only modeling a subset of possible human actions and human-object interactions, 2) the nature of the egocentric pose estimation task, 3) our network's incorporation of physical laws and scene context, which reduces the number of possible trajectories.

\vspace{-3mm}
\section{Universal Humanoid Controller}
\vspace{-1mm}
\label{app:uhc}
\subsection{Implementation Details}
\vspace{-1mm}
\noindent \textbf{Proxy humanoid.} The proxy humanoid we use for simulation is created automatically using the mesh, bone and kinematic tree defined in the popular SMPL \cite{Loper2015SMPLAS} human model. Similar to the procedure in \cite{yuan2021simpoe}, given the SMPL body vertices $V = 6890$ and bones $B = 25$, we generate our humanoid based on the skinning weight matrix $\bs W \in \mathbb{R}^{V \times B}$ that defines the association between each vertex and bone. The geometry of each bone's mesh is defined by the convex hull of all vertices assigned to the bone. The mass of each bone is in turn defined by the volume of the mesh. To simplify the simulation process, we discard all body shape information from the AMASS \cite{Mahmood2019AMASSAO} dataset, and use the mean body shape of the SMPL model. Since AMASS and our MoCap dataset are recorded by people with different height, we manually adjust the starting height of the MoCap pose to make sure each of the humanoid's feet are touching the ground at the starting point of the episode.

\noindent \textbf{Policy network architecture.} 
Our Universal Humanoid Controller (UHC)'s workflow and architecture can be seen in Fig. \ref{fig:uhc}. $\bs {\bs \pi}_{\text{UHC}}(\bs{a}_t|\bs{s}_t, \widehat{\bs{q}}_{t+1})$ is implemented as a multiplicative compositional policy (MCP) \cite{Peng2019MCPLC} with eight motion primitives, each being an MLP with two hidden layers (512, 256). The composer is another MLP with two hidden layers (300, 200) and outputs the multiplicative weights ${\bs w}^{1:n}_t$ for the n motion primitives. As studied in MCP \cite{Peng2019MCPLC}, this hierarchical control policy increases the model's capacity to learn multiple skills simultaneously. The output ${\bs a}_t \in \mathbb{R}^{75}$ is a vector concatenation of the target angles of the PD controller mounted on the 23 no-root joints (each has 3 DoF), plus the residual force \cite{yuan2020residual}: $\bs \eta_t \in \mathbb{R}^{6}$. Recall that each target pose $\widehat{\bs{q}}_t \in \mathbb{R}^{76}$, $\widehat{\bs{q}}_t  \triangleq (\widehat{\bs{r}}^{\text{pos}}_t, \widehat{\bs{r}}^{\text{rot}}_t, \widehat{\bs{j}}^{\text{rot}}_t)$ consists of the root position $\widehat{\bs{r}}^{\text{pos}}_t \in \mathbb{R}^{3}$, root orientation in quaternions $\widehat{\bs{r}}^{\text{rot}}_t \in \mathbb{R}^{4}$ , and body joint angles in Euler angles $\widehat{\bs{j}}^{\text{rot}}_t \in \mathbb{R}^{69}$ of the human model. The use of quaternions and Euler angles follows the specification of Mujoco \cite{Todorov2012MuJoCo}. As described in the main paper, our UHC first transforms the simulation state to a feature vector using ${\bs T}_{\text{AC}}\left({\bs{q}}_{t}, \dot{\bs{q}}_{t}, \widehat{\bs{q}}_{t+1}, D_{\text{diff}}(\widehat{\bs{q}}_{t+1}, {\bs{q}}_{t}) \right)$ to output a 640 dimensional vector that is a concatenation of the following values: 

\begin{figure}[t]
\centering
\includegraphics[width=\textwidth]{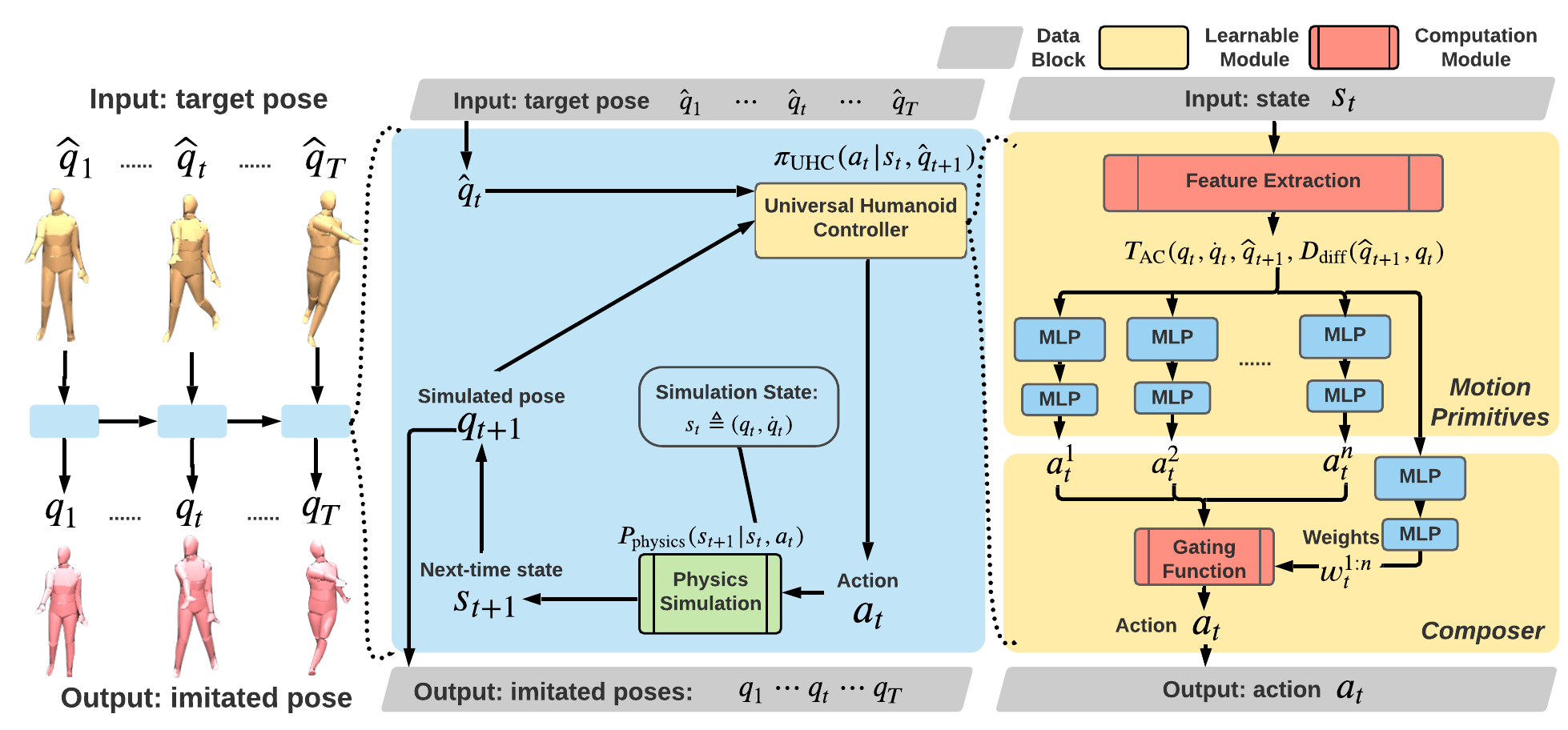}
\caption{Overview of our Universal Dynamics Controller. Given a frame of target pose and current simulation state, our UHC ${\bs \pi}_{\text{UHC}}$ can dirve the the humanoid to match the target pose.}
\label{fig:uhc}
\vspace{-3mm}
\end{figure}

\begin{equation}
\begin{aligned}
    ( {\bs h}^{\prime q}_t, {\bs q}^{\prime}_t, \widehat{\bs q}^{\prime}_t, ({\bs q}_t -  \widehat{\bs q}_t), \dot{\bs{q}}_{t}, ({\bs \psi}_t - \widehat{\bs \psi}_t),  &\widehat{\bs j}^{\prime \text{pos}}_t, ( {\bs j}^{\prime \text{pos}}_t -  \widehat{\bs j}^{\prime \text{pos}}_t), {\bs j}^{\prime \text{rot}}_t, ( {\bs j}^{\prime \text{rot}}_t \ominus  \widehat{\bs j}^{\prime \text{rot}}_t) ) \\ 
    &= {\bs T}_{\text{AC}}({\bs{q}}_{t}, \dot{\bs{q}}_{t}, \widehat{\bs{q}}_{t+1}, D_{\text{diff}}(\widehat{\bs{q}}_{t+1}, {\bs{q}}_{t}) ).
\end{aligned}
\end{equation}

It consists of:  root orientation ${\bs h}^{\prime q}_t \in \mathbb{R}^4$ in agent-centric coordinates;  simulated pose ${\bs q}^{\prime}_t \in  \mathbb{R}^{74}$ (${\bs q}^{\prime}_t \triangleq ({\bs r}^{\prime z}_t, {\bs r}^{\prime rot}_t, {\bs j}^{\text{rot}}_t) $, root height ${\bs r}^{\prime z}_t \in \mathbb{R}^{1}$, root orientation ${\bs r}^{\prime rot}_t \in \mathbb{R}^{4}$, and body pose ${\bs j}^{\text{rot}}_t \in \mathbb{R}^{69}$ expressed in Euler angles) in agent-centric coordinates; target pose $\widehat{\bs q}^{\prime}_t \in  \mathbb{R}^{74}$ in agent-centric coordinates; $({\bs q}_t -  \widehat{\bs q}_t) \in \mathbb{R}^{76}$ is the difference between the simulated and target pose (in world coordinate), calculated as  $({\bs q}_t -  \widehat{\bs q}_t) \triangleq (\widehat{\bs{r}}^{\text{pos}}_t - {\bs{r}}^{\text{pos}}_t, \widehat{\bs{r}}^{\text{rot}}_t \ominus {\bs{r}}^{\text{rot}}_t, \widehat{\bs{j}}^{\text{rot}}_t - {\bs{j}}^{\text{rot}}_t)$, where $\ominus$ calculates the rotation difference; $\dot{\bs{q}}_{t} \in \mathbb{R}^{75}$ is the joint velocity computed by Mujoco; $({\bs \psi}_t - \widehat{\bs \psi}_t) \in \mathbb{R}^{1}$ is the difference between the current heading (yaw) of the target and simulated root orientation; $ \widehat{\bs j}^{\prime \text{pos}}_t \in \mathbb{R}^{72}$ and $ ( {\bs j}^{\text{pos}}_t -  \widehat{\bs j}^{\text{pos}}_t) \in \mathbb{R}^{72}$ are joint position differences, calculated in the  agent-centric space, respectively;  $  \widehat{\bs j}^{\prime \text{rot}}_t \in \mathbb{R}^{96}$ and  $ ( {\bs j}^{\prime \text{rot}}_t \ominus  \widehat{\bs j}^{\prime \text{rot}}_t) \in \mathbb{R}^{96}$ are joint rotation differences in quaternions (we first convert $ \widehat{\bs j}^{\prime \text{rot}}_t$ from Euler angles to quaternions), calculated in the global and agent-centric space, respectively.

\noindent \textbf{Reward function.}
The imitation reward function per timestep, similar to the reward defined in Yuan \etal \cite{yuan2020residual} is as follows:
\begin{equation}
 r_t = w_{\text{jr}}  r_{\text{jr}} + w_{\text{jp}}  r_{\text{jp}} + w_{\text{jv}}  r_{\text{jv}} +   w_{\text{res}}  r_{\text{res}},
\end{equation}
where $w_{\text{jr}}, w_{\text{jp}}, w_{\text{jv}},  w_{\text{res}}$ are the weights of each reward. The joint rotation reward ${r_{\text{jr}}}$ measures the difference between the simulated joint rotation $\bs{j}^{\text{rot}}_t$ and the target $ \widehat{{\bs j}^{\text{rot}}_t}$ in quaternion for each joint on the humanoid. The joint position reward $\bs r_{\text{jp}}$ computes the distance between each joint's position $\bs j^{\text{pos}}_t$ and the target joint position $ \widehat{{\bs j}^{\text{pos}}_t}$. The joint velocity reward $\bs r_{\text{jv}}$ penalizes the deviation of the estimated joint angular velocity ${\dot{\bs j}}^{\text{rot}}_t$  from the target $\widehat{\dot{\bs j}}^{\text{rot}}_t$. The target velocity is computed from the data via finite difference. All above rewards include every joint on the humanoid model (including the root joint), and are calculated in the world coordinate frame. Finally, the residual force reward $\bs r_{\text{res}}$ encourages the policy to rely less on the external force and penalize for a large $\bs \eta_t$:

\begin{equation}
    \begin{aligned}
    r_{\text{jr}} = \exp\left [-2.0\left( \|{{\bs j}^{\text{rot}}}_t \ominus \widehat{\bs j}^{\text{rot}}_t \|^2\right)\right],& \quad 
    r_{\text{jp}} =\exp \left[-5\left(\left\|{\bs j}^{\text{pos}}_{t}-\widehat{\bs j}^{\text{pos}}_{t}\right\|^{2}\right)\right], \\
    r_{\text{jv}} =\exp [-0.005\left\|{\dot{\bs j}^{\text{rot}}}_t \ominus \widehat{\dot{\bs j}}^{\text{rot}}_t\right\|^{2}],& \quad 
    r_{\text{res}} =\exp \left[-\left(\left\| {\bs \eta}_t \right\|^{2}\right)\right].  \\  
    \end{aligned}
\end{equation}

We train our Universal Humanoid Controller for 10000 epoches, which takes about 5 days. Additional hyperparameters for training the UHC can be found in Table \ref{tab:hyper_uhc}:

\begin{table}[h]
\vspace{-5mm}
\caption{Hyperparameters used for training the Universal Humanoid Controller.} \label{tab:hyper_uhc}
\centering
\resizebox{\linewidth}{!}{%
\begin{tabular}{lcccccccc}
\toprule
  & $\gamma $& Batch Size & Value Learning Rate & Policy Learning Rate  &  PPO clip $\epsilon$ & Covariance Std
  \\ \midrule
Value   & 0.95   & 50000 & $3\times 10^{-4}$ & $5\times 10^{-5}$ & 0.2 & $0.1$
\\ \midrule
& $ w_{\text{jr}}$ & $ w_{\text{jp}}$  & $ w_{\text{jv}}$ &  $ w_{\text{res}}$ 
 & Sampling Temperature &  & \\ \midrule
Value   & 0.3    & 0.55    & 0.1    & 0.05    & 2    &  \\
\bottomrule 
\end{tabular}}\\ 
\vspace{-2mm}
\end{table}

\noindent \textbf{Training data clearning}
We use the AMASS \cite{Mahmood2019AMASSAO} dataset for training our UHC. The original AMASS dataset contains 13944 high-quality motion sequences, and around 2600 of them contain human-object interactions such as sitting on a chair, walking on a treadmill, and walking on a bench. Since AMASS does not contain object information, we can not faithfully recreate and simulate the human-object interactions. Thus, we use a combination of heuristics and visual inspection to remove these sequences. For instance, we detect sitting sequences through finding combinations of the humanoid's root, leg, and torso angles that correspond to the sitting posture; we find walking-on-a-bench sequences through detecting a prolonged airborne period; for sequences that are difficult to detect automatically, we conduct manual visual inspection. After the data cleaning process, we obtain 11299 motion sequences that do not contain human-object interaction for our UHC to learn from.

\subsection{Evaluation on AMASS}
\begin{wraptable}{r}{0.5\linewidth}
\vspace{-3mm}
\caption{Evaluation of motion imitation for our UHC using target motion from the AMASS dataset.}
\raggedleft
\resizebox{\linewidth}{!}{%
    \begin{tabular}[b]{lrrrrr}
    \toprule
    \multicolumn{5}{c}{AMASS dataset} 
    \\ 
    \midrule
    Method  & $\text{S}_{\text{inter}} \uparrow$ & $\text{E}_{\text{root}}\downarrow$ & $\text{E}_{\text{mpjpe}} \downarrow$   & $\text{E}_{\text{acc}} \downarrow$    \\ \midrule
    DeepMimic       & {24.0\%} &  {0.385} &  {61.634} &  {17.938}   \\ 
    UHC w/o MCP  &  {95.0\%} &  {0.134} &  25.254 &  {5.383} \\  
    UHC       &    \textbf{97.0\%} &  \textbf{0.133} &  \textbf{24.454} &  \textbf{4.460}  \\
    \bottomrule 
    \end{tabular}
    }\\ 
\label{tab:uhc_amass}
\end{wraptable}

To evaluate our Universal Humanoid Controller's ability to learn to imitate diverse human motion, we run our controller on the full AMASS dataset (after removing sequences that include human-object interactions) that we trained on. After data cleaning, the AMASS dataset contains 11299 high quality motion sequences, and contains challenging sequences such as kickboxing, dancing, backflipping, crawling, etc. We use a subset of metrics from egocentric pose estimation to evaluate the motion imitation results of UHC. Namely, we report $\text{S}_{\text{inter}}$, $\text{E}_{\text{root}}$, $\text{E}_{\text{mpjpe}}$, $\text{E}_{\text{acc}}$, where the human-object interation $\text{S}_{\text{inter}}$ indicates whether the humanoid has become unstable and falls down during the imitation process. The baseline we compare against is the popular motion imitation method DeepMimic \cite{Peng2018DeepMimic}. Since our framework uses a different physics simulation (Bullet \cite{coumans2021} vs Mujoco \cite{Todorov2012MuJoCo}), we use an in-house implementation of DeepMimic. From the result of Table \ref{tab:uhc_amass} we can see that our controller can imitate a large collection (10956/11299, 96.964\%) of realistic human motion with high fidelity without falling. Our UHC also achieves very low joint position error on motion imitation and, upon visual inspection, our controller can imitate highly dynamic motion sequences such as dancing and kickboxing. Failure cases include some of the more challenging sequences such as breakdancing and cartwheeling and can be found in the \href{https://zhengyiluo.github.io/projects/kin\_poly/}{supplementary video}.

\subsection{Evaluation on H36M}
To evaluate our Universal Humanoid Controller's ability to generalize to \textit{unseen motion sequences}, we use the popular Human 3.6M (H36M) dataset \cite{h36m_pami}. We first fit the SMPL body model to ground truth 3D keypoints similar to the process in \cite{Moon2020I2LMeshNetIP} and obtain motion sequences in SMPL parameters. Notice that this fitting process is imperfect and the resulting motion sequence is of less quality than original MoCap sequences. 
\begin{wraptable}{r}{0.5\linewidth}
\caption{Evaluation of motion imitation for our UHC using target motion from the H36M dataset.}
\raggedleft
\resizebox{\linewidth}{!}{%
    \begin{tabular}[b]{lrrrrr}
    \toprule
    \multicolumn{5}{c}{H36M dataset} 
    \\ 
    \midrule
    Method  & $\text{S}_{\text{inter}} \uparrow$ & $\text{E}_{\text{root}}\downarrow$ & $\text{E}_{\text{mpjpe}} \downarrow$   & $\text{E}_{\text{acc}} \downarrow$    \\ \midrule
    DeepMimic       & {0.0\%} &  {0.609} &  {107.895} &  {28.881}   \\ 
    UHC w/o MCP  &  {89.3\%} &  {0.200} &  \textbf{36.972} &  {4.723} \\  
    UHC       &    \textbf{92.0\%} &  \textbf{0.194} &  {40.424} &  \textbf{3.672}  \\
    \bottomrule 
    \end{tabular}
    }\\ 
\label{tab:uhc_h36m}
\end{wraptable} 
These sequences are also never seen by our UHC during training. As observed in Moon \etal \cite{Moon2020I2LMeshNetIP}, the fitted SMPL poses have a mean per joint position error of around 10mm. We use the train split of H36M (150 unique motion sequences) as the target pose for our UHC to mimic. From the results shown in Table \ref{tab:uhc_h36m}, we can see that our UHC can imitate the unseen motion in H36M with high accuracy and success rate, and outperforms the baseline method significantly. Upon visual inspection, we can see that the failure cases often result from losing balance while the humanoid is crouching down or starts running suddenly. Since our controller does not use any sequence level information, it has no way of knowing the upcoming speedup of the target motion and can result in instability. This indicates the importance of the kinematic policy adjusting its target pose based on the current simulation state to prevent the humanoid from falling down, and signifies that further investigation is needed to obtain a better controller. For visual inspection of motion imitation quality and failure cases, please refer to our supplementary video.

\vspace{-3mm}
\section{Additional Dataset Details}
\vspace{-1mm}
\label{app:data}

\begin{figure*}[b]
\begin{center}
\includegraphics[width=1\hsize]{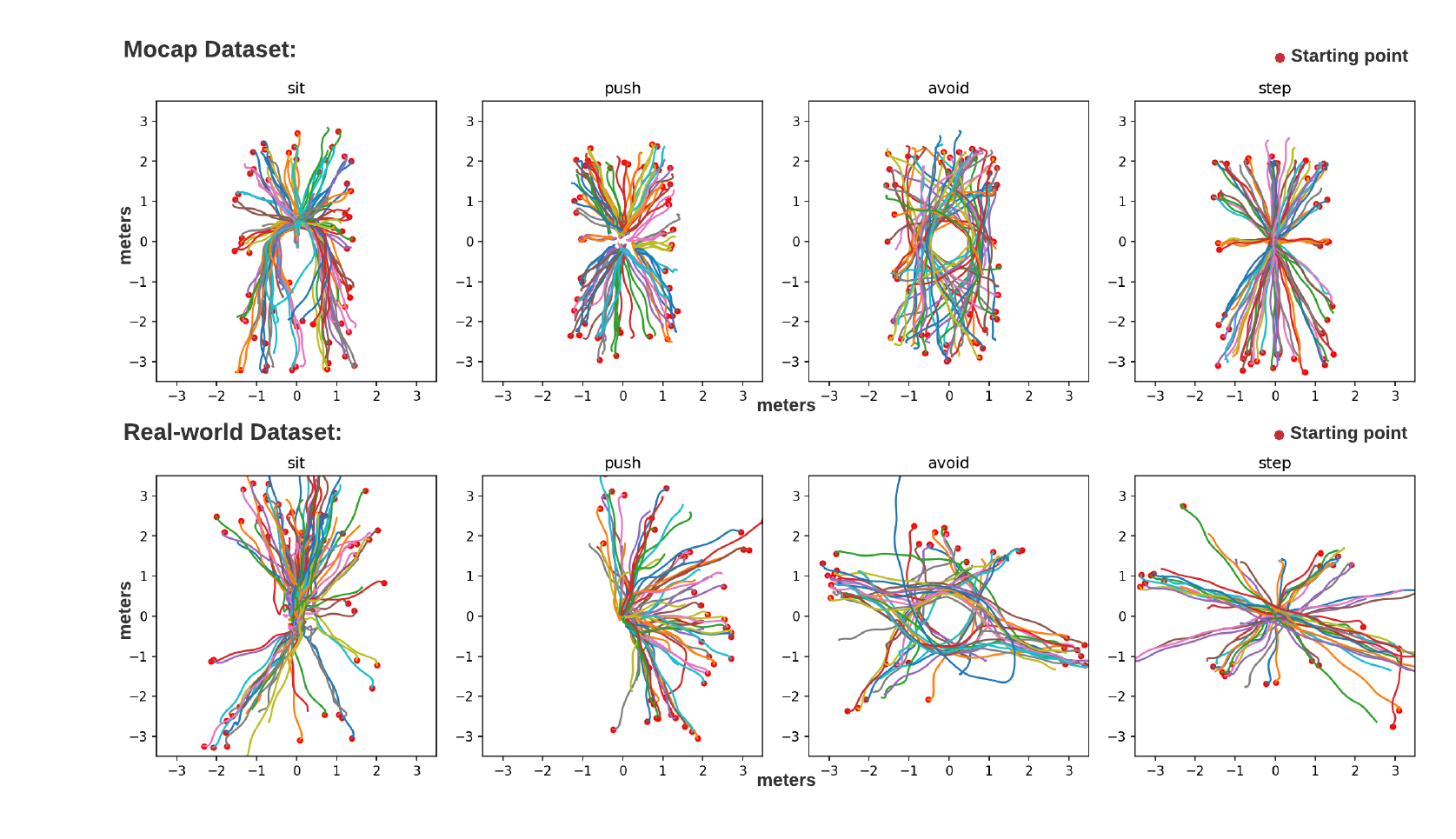}
\end{center}
\vspace{-2mm}
\caption{Trajectory analysis on our MoCap and real-world datasets. \textbf{Here we recenter each trajectory using the object position and plot the camera trajectory}, the objects are positioned differently across trajectories. The starting point is marked as a red dot. }
\vspace{-3mm}
\label{fig:mocapwild}
\end{figure*}

 \subsection{MoCap dataset.} 
 
 Our MoCap dataset (202 training sequences, 64 testing sequences, in total 148k frames) is captured in a MoCap studio with three different subjects. Each motion clip contains paired first-person footage of a person performing one of the five tasks: sitting down and (standing up from) a chair, avoiding an obstacle, stepping on a box, pushing a box, and generic locomotion (walking, running, crouching). Each action has around $50$ sequences. The locomotion part of our dataset is merged from the egocentric dataset from EgoPose \cite{Yuan_2018_ECCV} since the two datasets are captured using a compatible system. MoCap  markers are attached to the camera wearer and the objects to get the 3D full-body human pose and $6$DoF object pose. To diversify the way actions are performed, we instruct the actors to vary their performance for each action (varying starting position and facing gait, speed \etc). We followed the Institutional Review Board's guidelines and obtained approval for the collection of this dataset. To study the diversity of our MoCap dataset, we plot the trajectory taken by the actors in Fig. \ref{fig:mocapwild}. We can see that our trajectories are diverse and are spread out around a circle with varying distance from the objects. Table \ref{tab:data_speed} shows the speed statistics for our MoCap dataset.

\begin{wrapfigure}{r}{0.35\textwidth}
\vspace{-2mm}
  \begin{center}
    \includegraphics[width=0.35\textwidth]{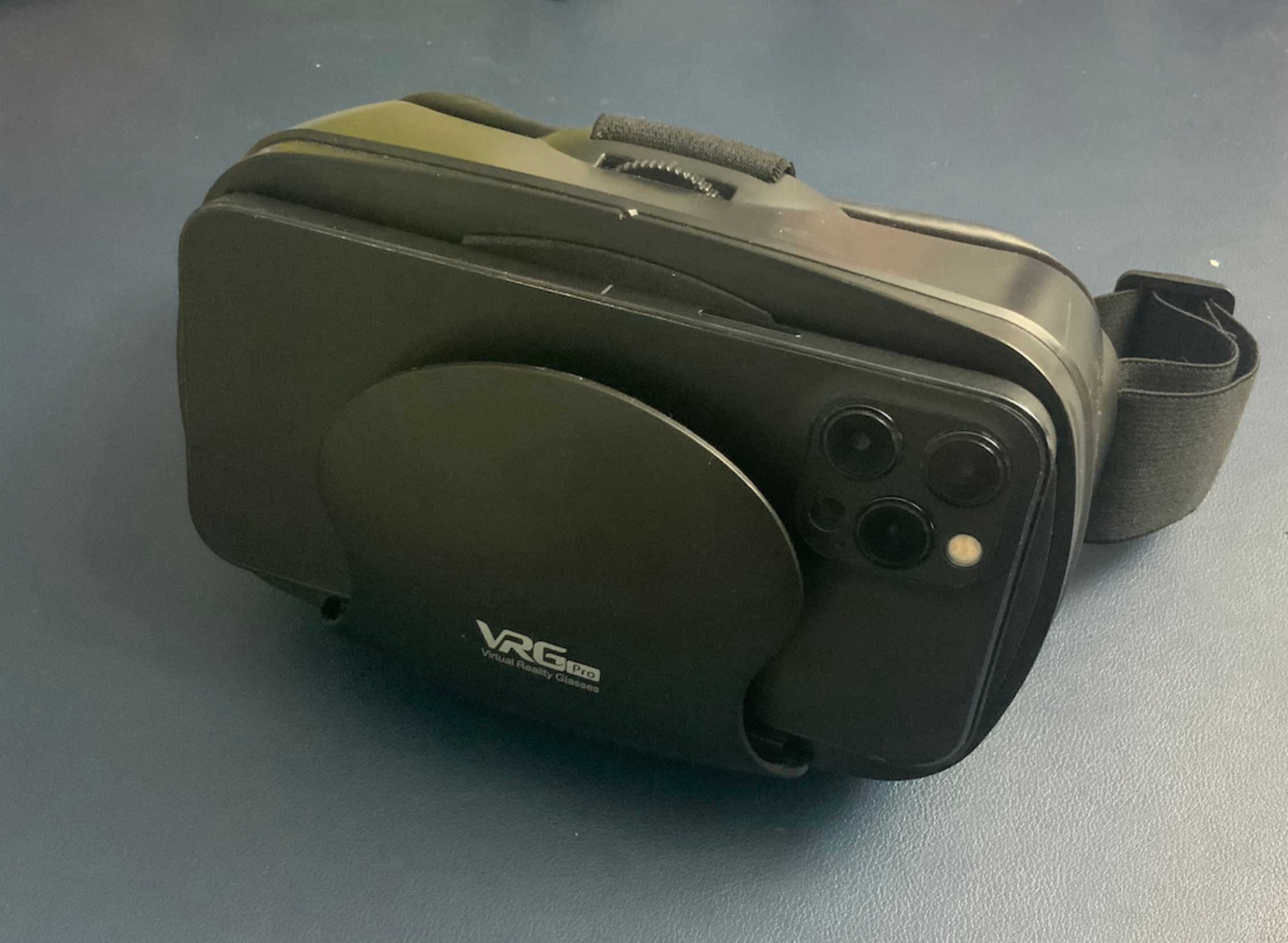}
  \end{center}
  \begin{footnotesize}
  \caption{
    Our real-world dataset capturing equipment.
  }\label{fig:head_iphone}
  \end{footnotesize}
  
  \vspace{-2mm}
\end{wrapfigure}

\subsection{Real-world dataset.} Our real world dataset (183 testing sequences, in total 55k frames) is captured in everyday settings (living room and hallway) with an additional subject. It contains the same four types of interactions as our MoCap dataset and is captured from a head-mounted iPhone using a VR headset (demonstrated in Fig.\ref{fig:head_iphone}).
Each action has around $40$ sequences. As can be seen in the camera trajectory in Fig. \ref{fig:mocapwild}, the real-world dataset is more heterogeneous than the MoCap dataset, and has more curves and banks overall. Speed analysis in Table \ref{tab:data_speed} also shows that our real-world dataset has a larger standard deviation in terms of walking velocity and has a larger overall spread than the MoCap dataset. In all, our real-world dataset has more diverse trajectories and motion patterns than our MoCap dataset, and our dynamics-regulated kinematic policy can still estimate the sequences recorded in this dataset. To make sure that our off-the-shelf object detector and pose estimator can correctly register the object-of-interest, we ask the subject to look at the object at the beginning of each capture session as a calibration period. Later, we remove this calibration period and directly use the detected object pose. 

Notice that our framework is starting \emph{position and orientation invariant}, since all of our input features are transformed into the agent-centric coordinate system using the transformation function $\bs{T}_{\text{AC}}$.

\begin{table}[t]
\caption{Speed analysis of our MoCap dataset and real-world dataset. Unit: (meters/second)} \label{tab:data_speed}
\centering
\begin{minipage}{.45\textwidth}
\begin{tabular}{lrrrr}
\toprule
\multicolumn{5}{c}{MoCap dataset } 
\\
\cmidrule{1-5} 
 Action & Mean  & Min & Max & Std
    \\ \midrule
Sit & {0.646}  & {0.442} & {0.837} & {0.098} \\  
 Push & {0.576} & {0.320} & {0.823}   & {0.119}\\  
 Avoid & {0.851}  & {0.567} & {1.084}  & {0.139}\\  
 Step & {0.844}  & {0.576} & {1.029} & {0.118} \\  
 
\bottomrule 
\end{tabular}\\ 
\end{minipage}
\begin{minipage}{.45\textwidth}
\begin{tabular}{lrrrr}
\toprule
\multicolumn{5}{c}{Real-world dataset } 
\\
\cmidrule{1-5} 
 Action & Mean  & Min & Max & Std
    \\ \midrule
 Sit & {0.556}  & {0.227} & {0.891}  & {0.171} \\  
 Push & {0.526}  & {0.234} & {0.762} & {0.127}  \\  
 Avoid & {0.668}  & {0.283} & {0.994} & {0.219} \\  
 Step & {0.729}  & {0.395} & {1.092} & {0.196} \\  
\bottomrule 
\end{tabular}\\ 
\end{minipage}
\vspace{-2mm}
\end{table}

\subsection{Dataset Diversity}

\section{Broader social impact.} Our overall framework can be used in extracting first-person camera wearer's physically-plausible motion and our humanoid controller can be a plug-and-play model for physics-based humanoid simulation, useful in the animation and gaming industry for creating physically realistic characters. There can be also negative impact from this work. Our humanoid controller can be used as a postprocessing tool to make computer generated human motion physically and visually realistic and be misused to create fake videos using Deepfake-like technology. Improved egocentric pose estimation capability can also mean additional privacy concerns for smart glasses and bodycam users, as the full-body pose can now be inferred from front-facing cameras only. As the realism of motion estimation and generation methods improves, we encourage future research in this direction to investigate more in detecting computer generated motion \cite{9412711}. 